\documentclass[lettersize,journal]{IEEEtran}  % Comment this line out if you need a4paper
\usepackage{graphics}    % for pdf, bitmapped graphics files
\usepackage{times}       % assumes new font selection scheme installed
\usepackage{amsmath}     % assumes amsmath package installed
\usepackage{amssymb}     % assumes amsmath package installed
\usepackage{graphicx}
\usepackage{algorithm}
\usepackage[noend]{algpseudocode}
\usepackage{subcaption}
\usepackage{threeparttable}
\usepackage{multirow}
\usepackage[hidelinks]{hyperref}
\usepackage{cite}
\usepackage{tikz}
\usepackage{xcolor}

\usetikzlibrary{ calc,
	positioning, % For easy node-relative placements
	patterns,
	decorations.pathreplacing,
	angles,quotes,intersections
}
\usepackage[font=small]{caption}

\def\secref#1{Sec.~\ref{#1}}
\def\figref#1{Fig.~\ref{#1}}
\def\tabref#1{Tab.~\ref{#1}}
\def\eqref#1{Eq.~(\ref{#1})}

\newcommand\etal{\emph{et al.}}

\def\argmin{\mathop{\rm argmin}}

\renewcommand{\d}[1]{\mbox{\boldmath$#1$}}	% vector style
\newcommand{\di}[1]{\mbox{\boldmath \hspace*{-0.01em}\scriptsize$#1$}}

\newcommand{\m}[1]{{\mbox{{\fontencoding{T1}\sffamily\slshape{#1\/}}}}}	% Matrix style
\newcommand{\mi}[1]{{\mbox{{\fontencoding{T1}\sffamily\slshape{\scriptsize#1\/}}}}}	
\newcommand{\trans}[0]{^{\sf T}}            % transpose
\algdef{SE}[DOWHILE]{Do}{doWhile}{\algorithmicdo}[1]{\algorithmicwhile\ #1}%

\algnewcommand{\LeftComment}[1]{\vspace{0.8ex}\Statex \(\triangleright\) {\bf #1}}
\algnewcommand{\Sort}[2]{{\bf Sort} {#1} {\bf by} {#2}}

\newcommand{\name}[0]{RDMNet}
\newcommand{\mname}[0]{{superpoint detection module}}
\newcommand{\PA}[0]{{\mathcal{P}^{\mi A}}}
\newcommand{\PB}[0]{{\mathcal{P}^{\mi B}}}

\newcommand{\A}[1]{{#1^{\mi A}}}
\newcommand{\B}[1]{{#1^{\mi B}}}
\newcommand{\mtf}[1]{{{\mathbf{#1}}}}
\newcommand{\hmtf}[1]{{{\hat{\mathbf{#1}}}}}

\newcommand{\mtc}[1]{{{{\mathcal{#1}}}}}

\usepackage{booktabs}

\makeatletter
\def\maketag@@@#1{\hbox{\m@th\normalfont\normalsize#1}}
\makeatother

\newcommand{\diff}[1]{\textcolor{black}{#1}}

\title{RDMNet: Reliable Dense Matching Based \\ Point Cloud Registration for Autonomous Driving}

\author{Chenghao Shi, Xieyuanli Chen, Huimin Lu$^*$, Wenbang Deng, Junhao Xiao$^*$, Bin Dai% <-this % stops a space
  \thanks{Chenghao Shi, Xieyuanli Chen, Huimin Lu, Wenbang Deng and Junhao Xiao are with the College of Intelligence Science and Technology, National University of Defense Technology, Changsha, China.}%
  \thanks{Bin Dai is with the Unmanned Systems Research Center, National Innovation Institution of Defense Technology, Beijing, China. }
  \thanks{Corresponding author: Junhao Xiao, Huimin Lu, e-mail: junhao.xiao@ieee.org, lhmnew@nudt.edu.cn. }
  \thanks{
  	This work was supported in part by the National Science Foundation of China under Grant U1913202, U22A2059 and U1813205, as well as
  	Major Project of Natural Science Foundation of Hunan Province under Grant 2021JC0004.
  }%
}

\begin{document}
\maketitle
%\thispagestyle{empty}
%\pagestyle{empty}
%\mark{IEEE TRANSACTIONS ON INTELLIGENT TRANSPORTATION SYSTEMS}
\markboth{}%
{Shi \MakeLowercase{\textit{et al.}}: RDMNet: Reliable Dense-point Matching for Robust and Accurate Point Cloud Registration}

%%%%%%%%%%%%%%%%%%%%%%%%%%%%%%%%%%%%%%%%%%%%%%%%%%%%%%%%%%%%%%%%%%%%%%%%%%%%%%%%

\begin{abstract}
Point cloud registration is an important task in robotics and autonomous driving to estimate the ego-motion of the vehicle. 	
Recent advances following the coarse-to-fine manner show promising potential in point cloud registration. 
However, existing methods rely on good superpoint correspondences, which are hard to be obtained reliably and efficiently, thus resulting in less robust and accurate point cloud registration. 
In this paper, we propose a novel network, named \name{}, to find dense point correspondences coarse-to-fine and improve final pose estimation based on such reliable correspondences. 
Our \name{} uses a devised 3D-RoFormer mechanism to first extract distinctive superpoints and generates reliable superpoints matches between two point clouds. The proposed 3D-RoFormer fuses 3D position information into the transformer network, efficiently exploiting point clouds' contextual and geometric information to generate robust superpoint correspondences. \name{} then propagates the sparse superpoints matches to dense point matches using the neighborhood information for accurate point cloud registration. 
We extensively evaluate our method on multiple datasets from different environments. \diff{The experimental results demonstrate that our method outperforms existing state-of-the-art approaches in all tested datasets with a strong generalization ability.}
\end{abstract}

%%%%%%%%%%%%%%%%%%%%%%%%%%%%%%%%%%%%%%%%%%%%%%%%%%%%%%%%%%%%%%%%%%%%%%%%%%%%%%%%

\section{Introduction}
\label{sec:intro}

Point cloud registration is a fundamental problem in computer vision, robotics, and autonomous driving. It aims to estimate the transformation between pairs of partially overlapped point clouds. 
The correspondence-based methods~\cite{huang2021cvpr,yu2021nips,bai2020cvpr} are the current domination. They first find the data association, such as point matches between two LiDAR point clouds. Based on that, they then compute the relative transformation straightforwardly with a singular value decomposition (SVD) or a robust estimator, e.g., RANSAC~\cite{fischler1981cacm}.
To balance the computation consumption and correspondence quality, most existing methods find the association on the downsampled sparse points or keypoints~\cite{bai2020cvpr, li2019iccv, huang2021cvpr}. 
However, downsampling will inevitably make part of the points lose their corresponding points, which degrades the registration performance.
\begin{figure}[t]
\vspace{-0.5cm}
	\centering
%	\DIFaddbeginFL \includegraphics[width=0.99\linewidth]{pics/motivation} \DIFaddendFL
	\includegraphics[width=0.99\linewidth]{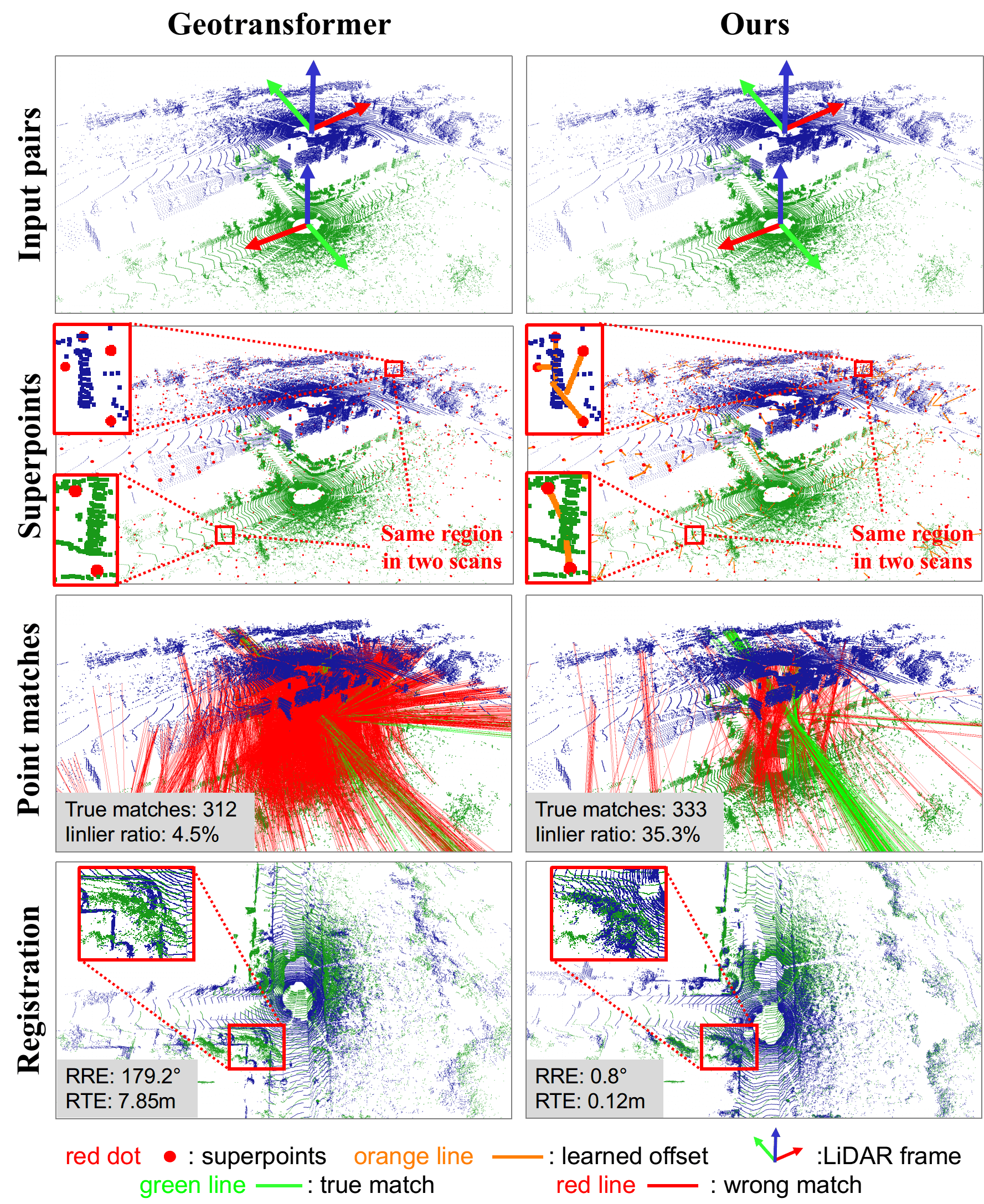}
	\caption{Point cloud registration under a challenging situation. We compare our method (right column) against GeoTransformer~\cite{zhu2022arxiv} (left column). Given two point clouds (first row), we first extract the sampled points uniformly distributed in the point cloud. Geotransformer uses them directly as superpoints in (second left), while our \name{} adds the learned offsets to the sampled points and generates better superpoints near the geometrically significant regions (second right). The orange lines show the learned offsets, which make superpoints from both point clouds fall closer in the geometrically significant regions. Using our proposed 3D-RoFormer to generate high-quality superpoint correspondences, our \name{} subsequently finds better dense-point correspondences (third right) compared to the baseline method (third left). In the end, our method successfully registers the two scans (bottom right) while the baseline method fails (bottom left). }
	\label{fig:motivation}
%	\vspace{-0.4cm}
\end{figure}

\diff{Inspired by works in image matching, recent advances~\cite{yu2021nips,qin2022cvpr} utilize the coarse-to-fine mechanism show remarkable performance on point cloud registration.
The coarse-to-fine mechanism~\cite{yu2021nips,qin2022cvpr} downsamples the point cloud into sparse superpoints and uses these superpoints to split the point cloud into point patches.} On the coarse level, it finds the superpoint (patch) correspondences depending on the overlapped area of two patches. \diff{On the fine level, the superpoint correspondences are then propagated to dense point matches based on neighborhood consensus, i.e., finding the point matches only from the matched patches.}
This limits the search space to a reasonable range, which not only reduces the computational complexity, but also makes the established matches more reliable. However, the performance of the final matches heavily relies on the quality of the superpoint correspondences.

In this paper, we dig into the properties of the superpoint that affect the final performance.
As a powerful contextual information encoder, transformer~\cite{vaswani2017nips} has been adapted to multiple point cloud learning tasks~\cite{yew2020cvpr}.  
Existing methods~\cite{yu2021nips,zhu2022arxiv,qin2022cvpr} exploit transformers to increase the robustness of superpoint matches.
However, the vanilla transformer used in CoFiNet~\cite{yu2021nips} lacks geometric information, which hinders the performance of the position-sensitive point cloud registration. GeoTransformer~\cite{zhu2022arxiv} infuses the pair-wise distance and triplet-wise angular information into the transformer, while NegNet~\cite{qin2022cvpr} constructs point pair features based on the geometric information~\cite{drost2010cvpr}.
Although yielding promising results, they are computationally expensive and neglect the distribution of the superpoints, thus leading to suboptimal point-matching results. 

To tackle the above-mentioned problems, we propose the \name{} to exploit both the contextual and geometric information of the point cloud and generate reliable dense-point correspondence for point cloud registration.
The core technique in \name{} is our devised novel transformation-invariant attention mechanism, named 3D-RoFormer.
It encodes the 3D position into a deep rotation matrix and naturally incorporates explicit relative position dependency into the self-attention calculation, thus becoming transformation-invariant while keeping lightweight and fast.
For the superpoint distribution, \name{} uses a \mname{} to first uniformly sample points over the whole point cloud and then learn the offset for each point, making the superpoint pairs more compact and falling in significant regions.
\figref{fig:motivation} demonstrates that our \name{} extracts more compact superpoint pairs and finds more reliable dense-point correspondences compared to the baseline methods.

To thoroughly evaluate our method, we conduct experiments on multiple outdoor datasets, including KITTI~\cite{geiger2012cvpr}, KITTI-360~\cite{liao2021arxiv}, Apollo~\cite{lu2019cvpr}, Mulran~\cite{zhang2021pr}, and a self-recorded dataset with our own mobile robot in a campus environment. Note that we only train our method on the training data of the KITTI dataset and directly apply it to other datasets collected by different LiDAR sensors in different environments. The experimental results show that our method outperforms the state-of-the-art methods in terms of both superpoint matching and pose estimation with strong generalization ability.

To sum up, our main contributions are:
\begin{itemize}
	\item A novel transformer, 3D-RoFormer, that efficiently learns the contextual and geometric features for superpoint matching with limited computation and storage cost.
	\item A novel network \name{} that generates reliable superpoints and dense-point correspondences to achieve state-of-the-art point cloud registration performance.
	\item Extensive evaluations on multiple outdoor datasets while only trained on the KITTI dataset show that our \name{} achieves superior performance with strong generalization ability compared to other state-of-the-art methods.
\end{itemize}
\diff{We will make the implementation of our method open-source.}

%%%%%%%%%%%%%%%%%%%%%%%%%%%%%%%%%%%%%%%%%%%%%%%%%%%%%%%%%%%%%%%%%%%%%%%%%%%%%%%%
\section{Related Work}
\label{sec:related}

Point cloud registration refers to finding the relative spatial transformation that aligns two point clouds. The existing methods can be broadly classified into two categories: correspondence-free and correspondence-based.

The correspondence-free methods transform the registration problem into a regression problem. Early work like PointNetLK proposed by Aoki \etal~\cite{aoki2019cvpr} first extracts the feature of the point cloud using PointNet~\cite{qi2017cvpr} and then regresses the transformation from the features. Zheng \etal~\cite{zheng2022tits} utilize a similar idea with PointNetLK and further refine the result in an iterative computation architecture. Other works such as the one by Huang \etal~\cite{huang2020cvpr} solve the registration problem by minimizing the feature-metric projection error. 
Such methods struggle to construct reliable regression models, and registration accuracy is not guaranteed.

The correspondence-based methods first extract correspondences between two point clouds and then compute the transformation using a direct solver or a robust estimator. Extracting correct correspondences is the most challenging part of such methods.
\diff{The standard correspondence-based approach is the iterative closest point~(ICP) algorithm~\cite{besl1992pami} and its numerous variants~\cite{zhang1994ijcv, segal2009rss}. They find correspondences using the nearest neighbor search or other heuristics iteratively, thus heavily relying on good initial estimation for transformation.
Recent work~\cite{yew2020cvpr} follows the idea of ICP and establishes the soft correspondences in the learned feature space, which relaxes the requirement of good initial guesses. However, the computational complexity and global searching limit the application of these methods to large-scale point clouds.}

\diff{Different from the above ICP-like methods, keypoint-based methods find correspondences on sparse points generated by either uniformly sampling~\cite{deng2018cvpr,deng2018eccv,li2021tits} or keypoint detection~\cite{yew2018eccv,li2019iccv, bai2020cvpr}.}
PPFNet and PPF-FoldNet proposed by Deng \etal~\cite{deng2018cvpr,deng2018eccv} introduce point pair feature~(PPF) combined with PointNet to produce local patch representation for matching.
Different from PPFNet and PPF-FoldNet establishing correspondences on uniform sampling points, keypoint-based methods sample points according to pre-defined~\cite{zhong2009iccvws} or learned saliency~\cite{yew2018eccv,li2019iccv, bai2020cvpr} for higher repeatability. 
\diff{In order to handle the low overlap situation, PREDATOR proposed by Huang \etal~\cite{huang2021cvpr} extracts the points that are not only salient but also lie in the overlap region. Based on that, Zhu \etal~\cite{zhu2022arxiv} recently proposed  NgeNet to augment the features with point pair information and use a multi-level consistency voting to improve discrimination.}

\begin{figure*}[t]
	\centering
	\includegraphics[width=0.99\linewidth]{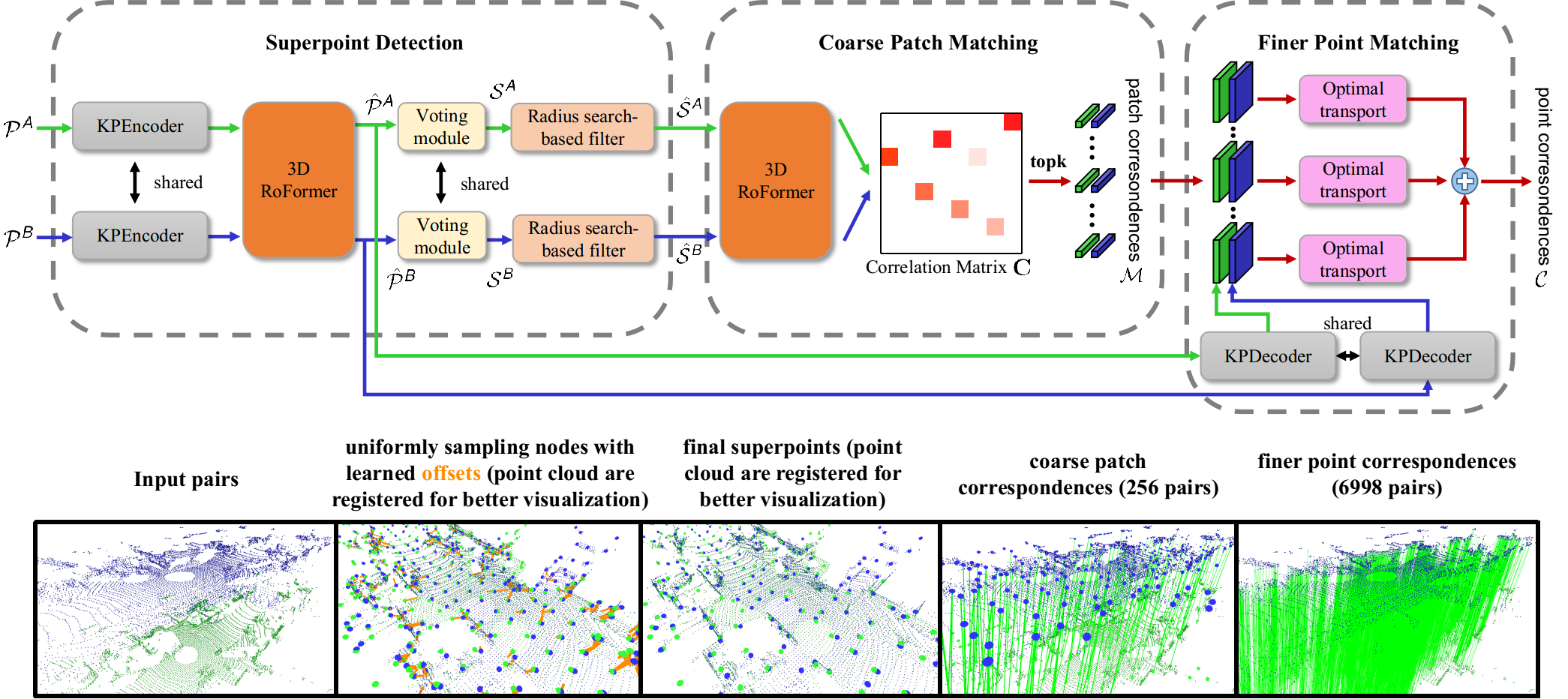}
	\caption{\textbf{Pipeline overview.} Given two point clouds, our \name{} first extracts superpoints from them using a superpoint detection module. Then, it applies a coarse patch matching to find the correspondences between sparse superpoints from two point clouds. Finally, a finer point matching module is then used to propagate superpoint correspondences into dense-point matches, which are used to estimate the final transformations between these two point clouds.} 
	\label{fig:model}
%	\vspace{-0.4cm}
\end{figure*}

Inspired by recent advances in image matching, 
CoFiNet proposed by Yu \etal~\cite{yu2021nips} and GeoTransformer proposed by Qin \etal~\cite{qin2022cvpr} utilize a coarse-to-fine mechanism that first finds reliable sparse point patch correspondences and then propagates the sparse correspondences to dense point matches.
Benefiting from the highly efficient backbone, they are able to obtain dense descriptors for large point clouds. The coarse-to-fine mechanism also greatly reduces the search space and increases the matching reliability. 
HRegNet proposed by Lu \etal~\cite{lu2021iccv} also utilizes the coarse-to-fine scheme, but different from the above methods, it extracts multi-level features and refines the transformation hierarchically. 
The quality of the sparse patch correspondences is important for such coarse-to-fine methods. To improve the correspondence accuracy, the existing methods mostly exploit the transformer network~\cite{vaswani2017nips}. 
For example, CoFiNet~\cite{yu2021nips} adopts the original transformer~\cite{vaswani2017nips} as a powerful contextual information encoder to generate more accurate point correspondences.
However, the vanilla transformer lacks geometric information, which hinders the performance of the position-sensitive point cloud registration. GeoTransformer~\cite{zhu2022arxiv} tackles this problem by infusing the pair-wise distance and triplet-wise angular into the transformer, while NegNet~\cite{qin2022cvpr} constructs the geometric features with point pair features~\cite{drost2010cvpr}.
Although yielding promising results, GeoTransformer results in extra-large $\mathcal{O}(n^2)$ storage complexity from the pair-wise distance and triplet-wise angular. NegNet is computationally expensive due to the normal estimation.
Besides, existing works neglect the distribution of the superpoints. They use simple uniform sampling points that may separate a single object into several patches, which usually retain a low overlap ratio with patches in the other point cloud. This may lead to bad superpoint matching and dense point propagation results.
Unlike the existing methods, our \name{} uses the proposed 3D-RoFormer exploiting both the contextual and geometric information of the point cloud, which generates better correspondences for point registration.

%%%%%%%%%%%%%%%%%%%%%%%%%%%%%%%%%%%%%%%%%%%%%%%%%%%%%%%%%%%%%%%%%%%%%%%%%%%%%%%%
\section{Our Approach}
\label{sec:main}
Given two point clouds $\PA=\{\d p_i^{\mi A} \in \mathbb{R}^3\}^M_{i=1}$ and $\PB=\{\d p_j^{\mi B} \in \mathbb{R}^3\}^N_{j=1}$, we aim to establish point correspondences between the two point clouds.
To this end, we propose the \name{} that finds correspondences in a coarse-to-fine manner. 
The overview of our approach is illustrated in \figref{fig:model}. It is built upon our devised novel 3D-RoFormer network (see~\secref{sec:3d-roformer}) and consists of three main steps: superpoint detection (see~\secref{sec:SD}), coarse patch matching (see~\secref{sec:coarse-level}), and finer point matching (see~\secref{sec:fine}). 
\vspace{-0.2cm}
\subsection{3D-RoFormer}
\label{sec:3d-roformer}
We first introduce our devised novel 3D-RoFormer, which is a translation-invariant transformer and the core technique of our \name{}.
We build 3D-RoFormer upon the vanilla transformer~\cite{vaswani2017nips}. 
For a point $\d p_i^{\mi Q}$ with its feature $\d h_i^{\mi Q}$ in the query point cloud $\m Q$ and all the points in the source point cloud $\m S$, the network computes the query~$\d q_i$, key~$\d k_j$, and value~$\d v_j$ transformer feature maps with a linear projection:
\begin{align}
\label{eq:attention}
\d q_i &= \m W_1 \, \d h_i^{\mi Q}+\d b_1,
\nonumber\\ 
\d k_j &= \m W_2 \, \d h_j^{\mi S}+\d b_2,
\\
\d v_j &= \m W_3 \, \d h_j^{\mi S}+\d b_3. \nonumber
\end{align} 

$\m Q,\m S$ could be downsampled input point clouds or sparse superpoints. If $\m Q,\m S$ are the same point cloud, \eqref{eq:attention} generates the feature maps for self-attention operation, otherwise cross-attention.
After that, the transformer computes an attentional weight for the query point with each source point: $\alpha_{ij}=\textrm{softmax}_j(\d q_i\trans\d k_j)$, and obtains the final attention-enhanced feature for the query point as:\diff{
\begin{align}
\label{eq:attention_output}
(\tilde{\d h}_{\text{vanilla}})_i = \sum_{j=1}^{|\mi{S}|}\textrm{softmax}_j(\d q_i\trans\d k_j) \d v_j =  \sum_{j=1}^{|\mi{S}|}\alpha_{ij} \d v_j,
\end{align} 
where $|\m{S}|$ represents the number of the points in $\m S$.}

Transformer has shown to be superior for point cloud registration~\cite{shi2021ral,qin2022cvpr,yu2021nips}. However, the vanilla transformer contains no geometric information, thus leading to suboptimal registration results. 
Different works are proposed to enhance the transformer with geometric information~\cite{qin2022cvpr,zhu2022arxiv}.
However, they are either memory-consuming~\cite{qin2022cvpr} or time-consuming~\cite{zhu2022arxiv}.

Inspired by the recent Roformer~\cite{su2021arxiv} using rotational information for natural language processing, we propose a novel 3D-Roformer that encodes the absolute position information with a rotation matrix for 3D point cloud registration. Based on 3D-Roformer, our \name{} can better exploit both contextual and geometric information of the point clouds to generate more reliable keypoints.
We first adapt the rotary position embedding into 3D data by leveraging a MLP and mapping the position $\hat{\d s}_i\in \mathbb{R}^3 $ into the rotary embedding $\d \Theta_i = [\theta_1, \theta_2, \cdots, \theta_{d/2}] \in \mathbb{R}^{ \frac{\hat{d}}{2}}$:
\begin{align}
\label{eq:mlp_angle}
\d \Theta_i = \text{MLP}_\text{rot}( \hat{\d s}_i).
\end{align}
Each element in $\d \Theta_i$ can be treated as a rotation in a 2D plane and represented by a rotation matrix. The final formulation of the rotary 3D position embeddings \mbox{$\m R_{ \d \scriptsize\Theta_i}\in\mathbb{R}^{\tilde{d}\times \tilde{d}}$} is:
\begin{small}
	\begin{align}
	\label{eq:rope_r}
	\m R_{ \d \scriptsize\Theta_i}
	&=
	\begin{bmatrix} &\cos \theta_1\!&-\sin \theta_1\!&\cdots\!&0\!&0
	\\ &\sin \theta_1\!&\cos \theta_1\!&\cdots\!&0\!&0
	\\&\vdots\!&\vdots\!&\ddots\!&\vdots\!&\vdots
	\\&0\!&0\!&\cdots\!&\cos \theta_{\frac{d}{2}}\!&-\sin \theta_{\frac{d}{2}}
	\\&0\!&0\!&\cdots\!&\sin \theta_{\frac{d}{2}}\!&\cos \theta_{\frac{d}{2}}
	\end{bmatrix}.
	% \nonumber \\
	%\d \Theta_i &=\{\theta_j,j\in[1,2,\cdots,d/2]\}.
	\end{align}
\end{small}

\begin{figure}[t] 
	\centering
	\includegraphics[width=0.99\linewidth]{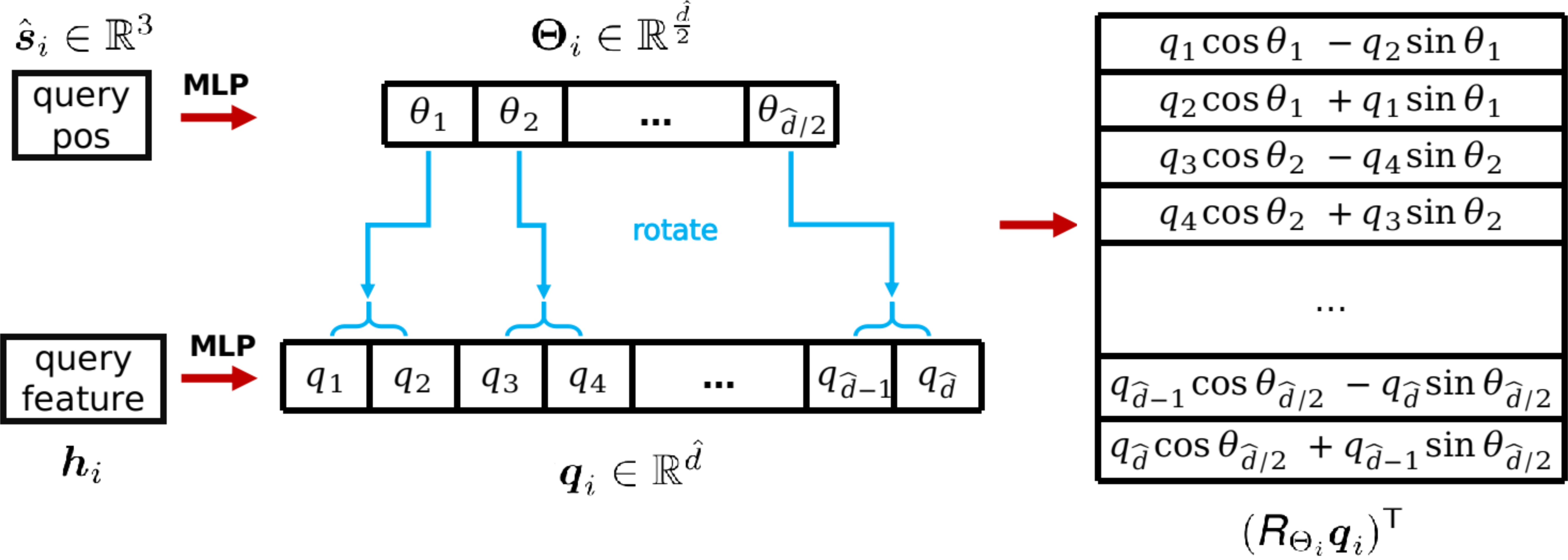}
	\caption{The rotary 3D position embedding.}
	\label{fig:rope} 
		\vspace{-0.2cm}
\end{figure}
Applying $\m R_{ \d \scriptsize\Theta_i}$ to a $\tilde{d}$ dimensional vector is equivalent to divide the vector into $\tilde{d}/2$ 2D vectors and rotate each of them by $\{\theta_i|i=1,\cdots,\tilde{d}/2\}$ accordingly (see \figref{fig:rope}).
We apply $\m R_{ \d \scriptsize\Theta_i}$ and $\m R_{ \d \scriptsize\Theta_j}$ to query~$\d q_i$ and key~$\d k_j$ respectively in self-attention operation and obtain the rotary self-attention as:
\begin{align}
\label{eq:roformer}
\alpha_{ij}''&=\textrm{softmax}_j((\m R_{ \d \scriptsize\Theta_i} \d q_i)\trans\m R_{ \d \scriptsize\Theta_j}\d k_j),\\
\tilde{\d h}_i
&=\sum_{j=1}^{|\tilde{\mathcal{P}}|}\alpha_{ij}''\d v_j. \label{eq:roformer3}
\end{align}

The benefits of using the proposed rotary self-attention are as follows. First, by encoding the position information as a rotation matrix, the rotary self-attention explicitly encodes the relative position information neatly.
Using the properties of the rotation matrix, we can further derive \eqref{eq:roformer} as:
\begin{align}
\label{eq:roformer2}
\alpha_{ij}''
&=\textrm{softmax}_j(\d q_i\trans\m R_{ \d \scriptsize\Theta_i}\trans\m R_{ \d \scriptsize\Theta_j}\d k_j),\nonumber\\
&=\textrm{softmax}_j(\d q_i\trans\m R_{ \d \scriptsize\Theta_j-\d \scriptsize\Theta_i}\d k_j),
\end{align}
where $\d \Theta_j-\d \Theta_i$ is naturally incorporated into the calculation of the attention scores $\alpha_{ij}''$ and then fused with the output feature $\tilde{\d h}_i$ in \eqref{eq:roformer3}. Therefore, our method is lightweight without requiring extra-large storage memory for relative position embeddings, which will be further verified in~\secref{sec:runtime}.

Secondly, the proposed 3D-Roformer is easy to deploy and operates very fast.
Due to the sparsity of $\m R_{ \d \scriptsize\Theta}$, the calculation of $\m R_{ \d \scriptsize\Theta_i}\cdot\d q_i$ and $\m R_{ \d \scriptsize\Theta_j}\cdot\d k_j$ can be done in a computationally efficient way using vector addition and multiplication operations, for example:
\begin{equation}
\small{
\label{eq:rope}
\m R_{ \d \scriptsize\Theta_i} \cdot \d q_i=
\begin{bmatrix} 
q_1\\q_2\\\vdots\\q_{d-1}\\q_{d}
\end{bmatrix}
\otimes
\begin{bmatrix} 
\cos\theta_1\\\cos\theta_1\\\vdots\\\cos\theta_{d/2}\\\cos\theta_{d/2}
\end{bmatrix}
+
\begin{bmatrix} 
-q_2\\q_1\\\vdots\\-q_{d}\\q_{d-1}
\end{bmatrix}
\otimes
\begin{bmatrix} 
\sin\theta_1\\\sin\theta_1\\\vdots\\\sin\theta_{d/2}\\\sin\theta_{d/2}
\end{bmatrix}.
}
% \nonumber \\
%\d \Theta_i &=\{\theta_j,j\in[1,2,\cdots,d/2]\},
\end{equation}

Thirdly, our proposed 3D-RoFormer is translation-invariant inherited from the linearity of MLP:
\begin{align}
	\label{eq:linearity}
	\d \Theta_j-\d \Theta_i = \text{MLP}_\text{rot}(\hat{\d {s}_j}-\hat{\d s}_i).
\end{align}
Therefore, our 3D-RoFormer will not be influenced by the changes in observation positions when used for finding correspondences. 
We enhance the final output features of the 3D-RoFormer $\tilde{\mathbf{H}}^{\mi A}$ and $\tilde{\mathbf{H}}^{\mi B}$ for point matching by interleaving the rotary self-attention and cross-attention for $l$ times.

%In practice, however, we find that using an activation function in \eqref{eq:mlp_angle} and bound the rotary embedding within 0 to $2\pi$ will result in a better performance:
%\begin{align}
%\label{eq:mlp_angle2}
%\d \Theta_i = 2\pi\cdot\sigma_\text{sigmoid}(\text{MLP}_\text{rot}( \hat{\d s}_i)),
%\end{align}
%though this makes the rotary self-attention no longer translation-invariant. The possible reason is that a non-linearity function can better represent the mapping from 3D position to rotary embedding.

Benefiting from the above-mentioned advantages, our proposed \name{} uses the devised 3D-RoFormer in both the superpoint detection and sparse patch matching modules to better find superpoint correspondences.
\vspace{-0.2cm}
\subsection{Superpoint Detection}
\label{sec:SD}
Our approach aims first to find reliable sparse patch matches and then propagate them to dense point matches based on neighborhood consensus. Such a coarse-to-fine scheme avoids the time-consuming and unreliable global search of dense feature correspondences.

Considering a point patch as the vicinity of a keypoint, the task can be treated as the keypoint detection and vicinity grouping.
We assign a keypoint for each point patch and regard them all together as one superpoint. A simple way to extract keypoints is to directly use the uniform sampling center point of each voxel~\cite{qin2022cvpr,yu2021nips}. 
However, the uniformly sampled center points may separate a single object into several patches and share low overlap vicinities with center points in the other point cloud (see~\figref{fig:motivation}), which leads to poor patch matching and dense point match propagation in the following steps.

To address this issue, we propose the \mname{} to extract more reliable superpoints for the point patch partition. 
We use the KPEncoder~\cite{thomas2019iccv} as the backbone of the proposed \mname{} that hierarchically downsamples and encodes the point cloud into the uniformly distributed nodes $\hat{\mathcal{P}}$ with associated features $\hat{\mathbf{F}}\in\mathbb{R}^{|\hat{\mathcal{P}}|\times \hat{d}}$.
The node feature from KPEncoder contains only the contextual and geometric information of the single point cloud but lacks information between two point clouds, thus not being able to reason the inter-clues to make the associated superpoints compact.
Therefore, we use the proposed 3D-RoFormer to fuse inter-point-cloud information and explicitly encode intra-point-cloud information at the same time. We denote the enhanced feature by 3D-RoFormer as $\tilde{\mathbf{F}}\in\mathbb{R}^{|\hat{\mathcal{P}}|\times \tilde{d}}$. 
A Voting module is then used to estimate the geometric offset and feature offset from the node to the proposal superpoint $\mtc{S}$, i.e., \mbox{$[\Delta{\mathbf{P}},\Delta{\mathbf{F}}]=\text{Vote}(\tilde{\mathbf{F}})$}, $\mtc{S}=\hat{\mathcal{P}}+\Delta\mathbf{P}$, and $\mtf{H}=\tilde{\mathbf{F}}+\Delta\mathbf{F}$.
We use a group of Multi-Layer Perceptron~(MLP) to form the Voting module. Though very simple, it generates meaningful offsets based on the feature learned by our 3D-RoFormer (see {\figref{fig:vote_vis}}) and boosts the registration performance by a large margin (see \tabref{tab:ablation}). 
In this paper, we supervise the superpoints to fall in the locally significant region, which may also lead to redundant proposals located in the same significant region. 
Thus, we use a simple radius search-based filtering strategy to force only one proposal in the same region. 
We iteratively perform a radius search for each proposal and filter out the ones close to the search center. After that, we obtain the final superpoints $\hat{\mathcal{S}}$ with associated features $\hat{\mathbf{H}}$.
Note that we limit the offsets $\Delta\mathbf{P}$ to a certain range, which maintains the superpoints to be evenly distributed throughout the point cloud instead of only concentrated in so-called significant areas. It also avoids possible degeneracy~\cite{li2019iccv}.

For each superpoint $\hat{\d s}_i$, we construct a local patch ${\mathcal{G}_i}$ using a point-to-node strategy~\cite{li2019iccv}. Specifically, each point is assigned to its nearest superpoint by:
\begin{align}
\label{eq:point-to-node}
{\mathcal{G}}_i = \{\d p\in\mathcal{P}|i=\argmin_j(\|{\d p}-\hat{\d s_j}\|_2),\hat{\d s_j}\in\hat{\mathcal{S}}\}.
\end{align} 

There are two advantages of this strategy. First, it assigns every point to a specific superpoint without duplication or loss. Second, it adapts to different densities, which is particularly suitable for our case since our superpoints break the uniformity of the original sampling after adding offsets.

\vspace{-0.2cm}
\subsection{Sparse Patch Matching}
\label{sec:coarse-level}
Based on the detected superpoints, we then conduct patch matching on the coarse level and find superpoints/patch correspondences between point clouds $\m A$ and $\m B$.

Since the superpoints have just been shifted and filtered by the \mname{}, the associated feature $\hat{\mathbf{H}}$ could be inconsistent with the surroundings.
Therefore, we first feed the superpoints with the associated features to another 3D-RoFormer to update the features with the newest contextual and geometric information from the updated superpoints.
Then we conduct the superpoint matching. We follow Qin~\etal~\cite{qin2022cvpr} and compute a Gaussian correlation matrix ${\mathbf{C}}\in\mathbb{R}^{|\hat{\mathcal{S}}^{\mi A}|\times|\hat{\mathcal{S}}^{\mi B}|}$ between normalized $\hat{\mathbf{H}}^{\mi A}$ and $\hat{\mathbf{H}}^{\mi B}$ with $c_{i,j}=\exp(-\|\A{\hmtf h_i}-\B{\hmtf h_j}\|^2)$. A dual-normalization is then performed to suppress ambiguous matches:
\begin{align}
\label{eq:dual_norm}
\hat{c}_{i,j}=\frac{c_{i,j}}{\sum_{k=1}^{|\A{\hat{\mathcal{S}}} |}c_{i,k}}\cdot\frac{c_{i,j}}{\sum_{k=1}^{|\B{\hat{\mathcal{S}}}|}c_{k,j}}.
\end{align}

We choose the largest $N_c$ entries as the superpoint correspondences:
\begin{align}
\label{eq:superpoint_corres}
{\mathcal{M}} = \{(\A{\hat{\d s}_{x_i}},\B{\hat{\d s}_{y_i}})|(x_i,y_i)\in\text{topk}_{x,y}(\hat{c}_{x,y})\}.
\end{align} 

Based on the fast superpoint matching, we determine the corresponding matched patches and use them as the basis for the subsequent fine-level dense-point matching.

\subsection{Dense Point Matching }
\label{sec:fine}
On the Fine level, we aim to generate dense point matches from the coarse patch correspondences. 

We leverage the KPDecoder~\cite{thomas2019iccv} to recover point-level descriptors $\mtf F$. 
Instead of using the updated superpoint feature, we recover the point-level feature from the raw anchor point feature $\tilde{\mtf F}$ since there might be information loss after offsetting and filtering.
\diff{For each superpoint correspondence $(\A{\hat{\d s}_{x_i}},\B{\hat{\d s}_{y_i}})$, we have its corresponding patch match $(\A{\mtc G_{x_i}}, \B{\mtc G_{y_i}})$ and then compute a match score matrix ${\mathbf{O}_i}\in\mathbb{R}^{M_i\times N_i}$ :
\begin{align}
\label{eq:OT}
\mathbf{O}_i = \A{\mtf{F}_{x_i}}(\B{\mtf{F}_{y_i}})\trans/\sqrt{\tilde{d}},
\end{align} 
where $M_i=|{\mathcal{G}}_{x_i}^{\mi A}|$ and $N_i=|{\mathcal{G}}_{y_i}^{\mi B}|$ represent the number of points in ${\mathcal{G}}_{x_i}^{\mi A}$ and ${\mathcal{G}}_{y_i}^{\mi B}$ respectively.}

To handle non-matched points, we append a ``dustbin" row and column for $\mathbf{O}_i$ filled with a learnable parameter $\alpha\in\mathbb{R}$. \diff{The Sinkhorn algorithm is then used to solve the soft assignment matrix $\mtf{Z}^i\in\mathbb{R}^{(M_i+1)\times(N_i+1 )}$.} Different from~\cite{qin2022cvpr,yu2021nips} that drops the dustbin and recovers the assignment by comparing the soft assignment score with a hand-tuned threshold, we directly find max entry both row-wise and column-wise on $\mtf{Z}^i$ which is then recovered to assignment $\mtc{C}^i$:

\begin{footnotesize}
	\begin{align}
	\label{eq:dense_match}
	\mtc{C}^i
	=&\{(\A{\mtc G_{x_i}}(m),\B{\mtc G_{y_i}}(n)|(m,n)\in \text{toprow}_{m,n}(\mtf{Z}_{1:M_i,1:(N_i+1)}^i)\}\cup\nonumber\\
	&\{(\A{\mtc G_{x_i}}(m),\B{\mtc G_{y_i}}(n)|(m,n)\in \text{topcolumn}_{m,n}(\mtf{Z}_{1:(M_i+1),1:N_i}^i)\},
	\end{align} 
\end{footnotesize}\diff{
where $m$ and $n$ represent the indexes of the max entry of $\mtf{Z}^i$.}

A point is either assigned to points in the matched patch or to the dustbin.
By this, we do not need manual tuning but require a discriminative assignment matrix, which can be obtained by using our proposed loss function as detailed in \secref{sec:loss}.
Note that a point is not strictly assigned to a single point in our approach, as the strict one-to-one point correspondences do not hold in practice due to the sparsity nature of the point cloud. 
Instead, we trust and keep the assignment results from both sides, i.e., matches from query to source and vice versa.
This results in extensively more point matches while maintaining a high inlier ratio, which benefits the transformation estimation.
The final correspondences are the combination of points matches from all patches:
\begin{align} 
	\mtc{C}=\bigcup_{i=1}^{N_c}\mtc{C}^i.
\end{align}

\subsection{Loss function and training}
\label{sec:loss}
\diff{The final loss is the weighted sum of these three components: \mbox{$L =  L_\text{s}+L_\text{c}+ L_\text{f}$}, where $L_\text{s}$ is the  superpoint detection loss, $L_\text{c}$ is the coarse match loss, $L_\text{f}$ is the fine match loss.}

\textbf{Superpoint detection loss}. The superpoint detection loss is composed of two parts $L_{s}=L_{s1}+L_{s2}$. The first part $L_{s1}$ is designed to guide our superpoints lying in the significant region, and the second part $L_{s2}$ is designed to make the superpoints close to the real measurement points.
Specifically, for the first part, we do not explicitly define the significance of a point, but use a chamfer loss to minimize the distance between matched superpoints:

\begin{equation}
\small{
\label{eq:ls1}
L_{\text{s}1}=\sum_{i=1}^{|\mtc{S}^{\mi A}|}\min_{\di s^{\mi B}_j\in\mtc{S}^{\mi B}}\|\d s^{\mi A}_i-\d s_j^{\mi B}\|^2_2+\sum_{i=1}^{|\mtc{S}^{\mi B}|}\min_{\di s_j^{\mi A}\in\mtc{S}^{\mi A}}\|\d s_i^{\mi B}-\d s_j^{\mi A}\|^2_2.
}
\end{equation}

Supervised by $L_{\text{s}1}$, we find that the superpoints tend to move to their nearest ``significant" regions to minimize the distance between superpoint pairs.
For the second part, we use another chamfer loss that minimizes the distance between the superpoint to its closest point:
\begin{equation}
\small{
\label{eq:ls2}
L_{s2}=\sum_{i=1}^{|\mtc{S}^{\mi A}|}\min_{\di p^{\mi A}_j\in\mtc{P}^{\mi A}}\|\d s^{\mi A}_i-\d p_j^{\mi A}\|^2_2+\sum_{i=1}^{|\mtc{S}^{\mi B}|}\min_{\di p_j^{\mi B}\in\mtc{P}^{\mi B}}\|\d s_i^{\mi B}-\d p_j^{\mi B}\|^2_2.
}
\end{equation} 

\textbf{Coarse match loss}. We follow~\cite{qin2022cvpr} and use overlap-aware circle loss to guide the network to extract reliable superpoint correspondence with relatively high overlap. 
We set the patch in $\m A$ with at least one positive patch in $\m B$ as the anchor patches $\A{\mtc{G}}$. 
For an anchor patch $\A{\mtc{G}_i}$, its positive patch set $\varepsilon_i^+$ is defined as those sharing at least 10\% overlap with $\A{\mtc{G}_i}$, and its negative patch set $\varepsilon_i^-$ is those that do not overlap with $\A{\mtc{G}_i}$. Then the overlap-aware circle loss on $\m A$ is calculated as: 
\begin{equation}
\footnotesize{
	\label{eq:coarse}
	\A{{L}_{\text{c}}} = 
\frac{1}{|\mtc{A}|}
	\sum_{\A{ \mtc{G}_i} \in\mtc{A}}
	\log[1+\hspace{-0.1cm}
	\sum_{\B{ \mtc{G}_j}\in\varepsilon^+_i}e^{\lambda_i^j\beta_{i,j}^+(d_i^j-\Delta^+)}
	\cdot\hspace{-0.1cm}
	\sum_{\B{\mtc{G}_k}\in\varepsilon^-_i}e^{\beta_{i,k}^-(\Delta^--d_i^k)}],
}
\end{equation}
where $d_i^j=\|\A{\hmtf h_i}-\B{\hmtf h_j}\|_2$, $\lambda_i^j$ refers to the overlap ratio between $\A{\mtc{G}_i}$ and $\B{\mtc{G}_j}$, and $\beta_{i,j}^+=\gamma(d_i^j-\Delta^+)$ and $\beta_{i,k}^-=\gamma(d_i^k-\Delta^-)$ represent the positive and negative weights. The hyper-parameters setting is followed by convention: $\Delta^+=0.1$ and $\Delta^-=1.4$.
The overall coarse match loss is the average of overlap-aware circle loss on $\m A$ and $\m B$, i.e., $L_{\text{c}}=(\A{{L}_{\text{c}}}+\B{{L}_{\text{c}}})/2$

\textbf{Fine match loss}. To learn a discriminative soft assignment matrix and support our dense point match module, we use a gap loss on the soft assignment matrix $\mtf{Z}^{i}$ of each patch correspondence $\{\A{\mtc G_{x_i}},\B{\mtc G_{y_i}}\}$.
For each matched patch pair, we generate its ground truth correspondences \mbox{$\mtf{M}^i \in \{0,1\}^{(M_i{+}1)\times(N_i{+}1)}$} with a match threshold $\tau$, where $M_i=|\A{\mtc G_{x_i}}|$, $N_i=|\B{\mtc G_{x_i}}|$.
The gap loss is then calculated as: 
\begin{align}
\label{eq:gap}
L_{\text{f}}^i =&\frac{1}{M_i}\sum_{m=1}^{M_i}\log(\sum_{n=1}^{N_i+1}[(- r_{m}^i+\mtf{Z}_{m,n}^i+\eta)_+ +1])
\nonumber\\ &+\frac{1}{N_i}\sum_{n=1}^{N_i}\log(\sum_{m=1}^{M_i+1}[(- c_{n}^i+\mtf{Z}_{m,n}^i+\eta)_+ +1]),
\end{align}
where $(\bullet)_+=max(\bullet,0)$, $r_{m}^i = \sum_{n=1}^{N_i{+}1}\mtf{Z}_{m,n}^i\mtf M_{m,n}^i$ refers to the soft assignment value for the true match of $m$-th point in $\A{\mtc G_{x_i}}$, and $c_{n}^i = \sum_{m=1}^{M_i{+}1}\mtf{Z}_{m,n}^i\mtf M_{m,n}^i$ refers to the soft assignment value for the true match of $n$-th point in $\B{\mtc G_{x_i}}$.
The final fine match loss is the average over all the matched patch pairs: \mbox{$L_\text{f}=\frac{1}{2|\mtc{M}|}\sum_{i=1}^{|\mtc{M}|}L_{\text{f}}^i$}.
 
\diff{We implement and evaluate our \name{} on 4 NVIDIA RTX 3090 GPUs. The network is trained with Adam optimizer~\cite{kingma2014arxiv}. We use 5 layers of KPEncoder~(4 layers of downsampling) and 3 layers of KPDecoder, which result in coarse-level points with a resolution of $4.8$~m and fine-level points with a resolution of $0.6$~m. The batch size is 1, and the learning rate is $10^{-4}$ and decay exponentially by 0.05 every 4 epochs. We also adapt the same data augmentation as in~\cite{huang2021cvpr}.}
 
%%%%%%%%%%%%%%%%%%%%%%%%%%%%%%%%%%%%%%%%%%%%%%%%%%%%%%%%%%%%%%%%%%%%%%%%%%%%%%%%

\section{Experimental Evaluation}
\label{sec:exp}
%Our experiments are designed to show the capabilities of our method.%

\begin{table*}
	\vspace{-0.2cm}
	\caption{Matching results on multiple datasets under different numbers of samples. The best results are highlighted in bold, and the second bests are marked with an underline.}
	\centering
	\scriptsize
	\setlength\tabcolsep{6pt}
%	\footnotesize
	\begin{tabular}{l|ccc|ccc|ccc|ccc|ccc}
		\toprule
		&\multicolumn{3}{c|}{KITTI}     &\multicolumn{3}{c|}{KITTI-360}   &\multicolumn{3}{c|}{Apollo}  		&\multicolumn{3}{c|}{Mulran} 	&\multicolumn{3}{c}{Campus}\\
%		\# Number of pairs	& \multicolumn{3}{c|}{555}  		&\multicolumn{3}{c|}{6238} 		&\multicolumn{3}{c|}{4250} 			&\multicolumn{3}{c|}{3363}		&\multicolumn{3}{c}{171}	\\
		\midrule
%		\multirow{2}*{Sensors} &\multirow{2}*{Velodyne 64}  &\multirow{2}*{Velodyne 64}  &\multirow{2}*{Velodyne 64} & Ouster 64	&\multirow{2}*{Velodyne 16}	\\
%		&   &   &  & (partially occluded)	&		\\
		
%		Sensors	&\multicolumn{3}{c|}{Velodyne 64}  		&\multicolumn{3}{c|}{Velodyne 64} 		&\multicolumn{3}{c|}{Velodyne 64} 			&\multicolumn{3}{c|}{Ouster 64}		&\multicolumn{3}{c}{Velodyne 16}	\\
		
		\# Samples &5000&1000&250     &5000&1000&250   &5000&1000&250  		&5000 &1000&250 	&2000&1000&250\\
		
		\midrule
		\multicolumn{16}{c}{\textit{Feature Match Recall}~(\%)}\\
		\midrule
		Predator~\cite{huang2021cvpr}	
		&{99.64}&{99.64}&{99.64} 		&99.87 &99.78&99.44			&{99.98}&{99.98}&{97.86} 	&{85.96}&82.01&58.76 	& 49.12&49.12&39.18\\ 
		CofiNet~\cite{yu2021nips}		
		&{99.64}&{99.64}&\textbf{99.82}		 &99.89&99.89&\underline{99.86} 			&\textbf{100}&\textbf{100}&\textbf{100}  		&91.79 &93.10&93.20	&62.57&66.07&67.84 \\ 
		NgeNet~\cite{zhu2022arxiv}		
		&{99.64}&{99.64}&{99.64}    	&\textbf{99.94}&\textbf{99.94}&\textbf{99.94}  	&\textbf{100}&\textbf{100}&\textbf{100}  		&\underline{95.18}&\underline{94.90}&87.70  &91.81&88.89&70.76\\ 
		Geotransformer~\cite{qin2022cvpr}
		&\textbf{99.82}&\textbf{99.82}&\textbf{99.82} 		&99.89&\underline{99.92}&\textbf{99.94}		&\textbf{100}&\textbf{100}&\textbf{100}  		&88.28&91.05&\underline{91.67} &\underline{98.25}&\underline{98.25}&\underline{97.66}\\
		\name~(\textit{ours})	
		&\textbf{99.82}&\textbf{99.82}&\textbf{99.82} &\underline{99.92}&\textbf{99.94}&\textbf{99.94}   &\textbf{100}&\textbf{100}&\textbf{100}  &\textbf{98.31}&\textbf{98.72}&\textbf{98.81}  &\textbf{100}&\textbf{100}&\textbf{100}  \\

		\midrule
		\multicolumn{16}{c}{\textit{Inlier Ratio}~(\%)}\\
		\midrule
		Predator~\cite{huang2021cvpr}	
		&62.9&50.2&29.5		&60.2&48.8&29.3			&44.3&31.8&16.8 		&14.2&11.1&6.6 		&5.5&5.4&4.7\\ 
		CofiNet~\cite{yu2021nips}		
		&34.2&36.1&36.2		&32.4&34.4&34.8 		&40.4&41.9&42.2 		&17.6&19.1&19.4 	&6.5&6.7&6.8\\ 
		NgeNet~\cite{zhu2022arxiv}		
		&66.5&{51.5}&28.6	    &63.2&49.6&28.5		&68.6&49.2&24.8 		&{29.1}&20.7&11.1  		&12.4&11.4&8.1\\ 
		Geotransformer~\cite{qin2022cvpr}
		&\underline{75.7}&\underline{86.0}&\underline{87.5} 	&\underline{73.2}&\underline{83.7}&\underline{85.5}			&\underline{83.8}&\underline{91.0}&\underline{92.4}		&\textbf{33.6} &\textbf{43.7}&\underline{46.3}	&\underline{19.0}&\underline{21.7}&\underline{24.9}\\
		\name~(\textit{ours})	
		&\textbf{86.7}&\textbf{93.0}&\textbf{95.3} &\textbf{84.0}&\textbf{91.0}&\textbf{93.7}   &\textbf{92.1}&\textbf{96.4}&\textbf{97.6}   &\underline{31.4}&\underline{42.6}&\textbf{51.1} 	&\textbf{34.9}&\textbf{36.8}&\textbf{41.5}  \\
		
		\bottomrule	
	\end{tabular}
	
		\vspace{-0.2cm}
	\label{tab:correspondences}
\end{table*}

\subsection{Dataset Overview}
\begin{figure}[t] 
	\centering
%	\DIFaddbeginFL \includegraphics[width=0.9\linewidth]{pics/platform} \DIFaddendFL
	\includegraphics[width=0.9\linewidth]{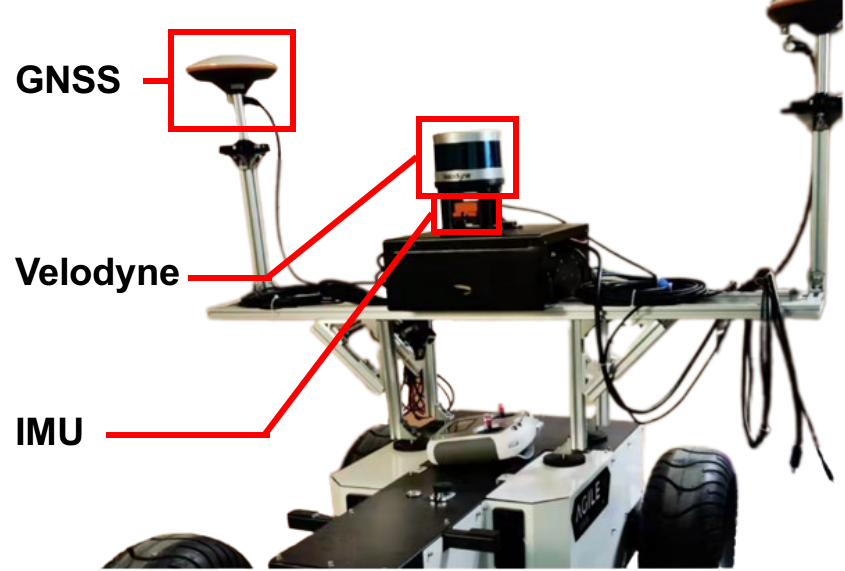}
	\includegraphics[width=0.9\linewidth]{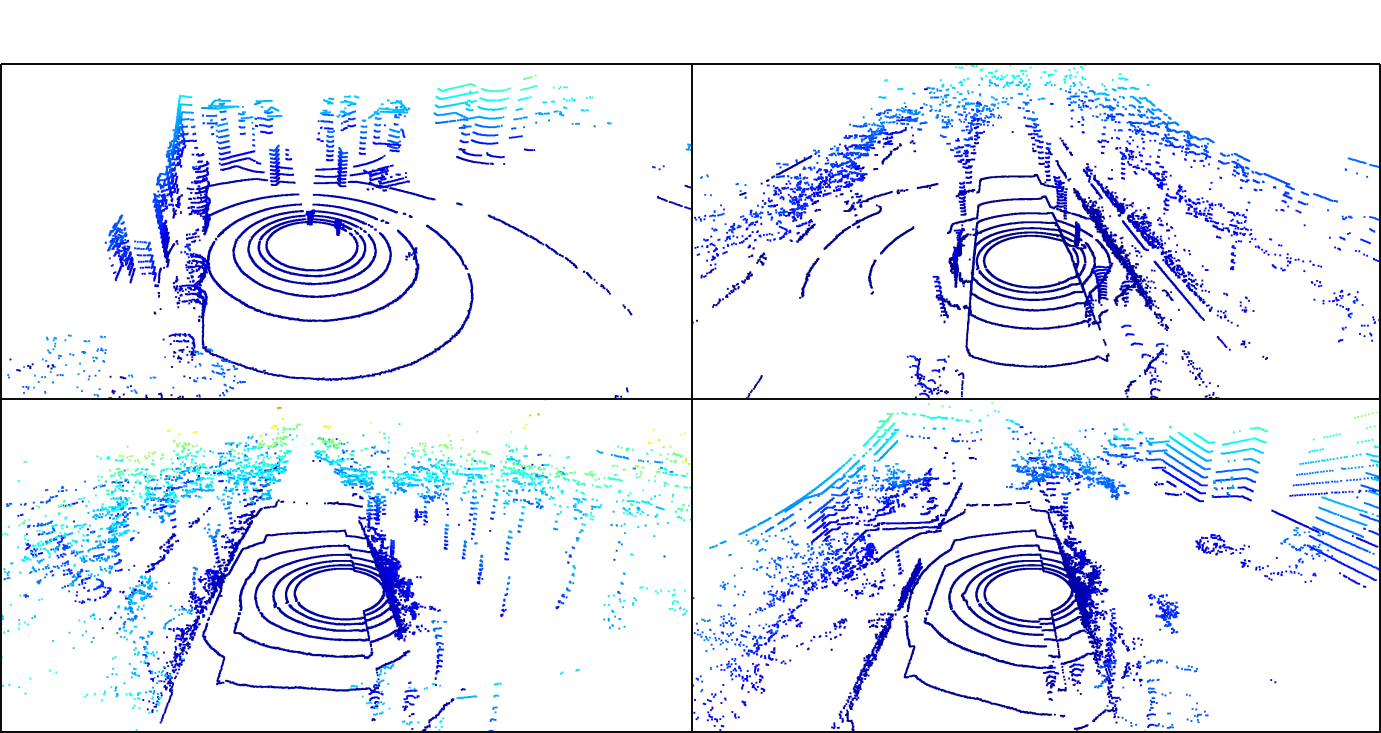}
	\caption{Our campus data collection platform and some LiDAR data visualization of the campus dataset.}
	\label{fig:platform} 
		\vspace{-0.3cm}

\end{figure}
%\begin{figure}[t]
%	\centering
%	\subcaptionbox[platform]{
%		\includegraphics[width=0.9\linewidth]{pics/platform}} %%width调整图片大小
%%	\hspace{10mm} %% 间隙
%	\subcaptionbox[dataset]{
%		\includegraphics[width=0.9\linewidth]{pics/campus_vis}}
%%	\caption{11}
%
%\end{figure}

We evaluate \name{} and compare it with the state-of-the-art methods on both publicly available datasets, including KITTI odometry~\cite{geiger2012cvpr}, KITTI-360~\cite{liao2021arxiv}, Apollo-SouthBay~\cite{lu2019cvpr} and Mulran~\cite{zhang2021pr} datasets, and a self-recorded dataset. These datasets provide LiDAR scans collected in different environments with the corresponding ground-truth poses. The KITTI odometry and KITTI-360 contain LiDAR data collected by a Velodyne HDL64 LiDAR in Germany. These two datasets use a similar sensor setup but collect data from different times and environments. The Apollo-SouthBay dataset also uses a Velodyne HDL64 LiDAR but with a different sensor setup collecting data in the U.S. cities. The Mulran dataset contains data collected by an OS1-64 LiDAR from Korea. \figref{fig:platform} shows our own platform equipped with a Velodyne VLP16 LiDAR, an inertial measurement unit~(Xsens MTi-300), and a GNSS~(INS CGI-410). We build our own dataset in a campus environment with ground-truth poses calculated by combining the GNSS and IMU with the state-of-the-art LiDAR SLAM method~\cite{shan2020Iiros}.

We follow~\cite{ huang2021cvpr, qin2022cvpr} and split the KITTI odometry into three sets: sequences 00-05 for training, 06-07 for validation, and 08-10 for testing. 
To evaluate the generalization ability, we directly apply the models trained on the KITTI odometry dataset to other datasets.
Also in line with~\cite{ huang2021cvpr, qin2022cvpr, zhu2022arxiv}, we use the LiDAR pairs that are at most 10~m away as samples and get 1358 pairs for training, 180 pairs for validation, and 14577 pairs for testing. 
Note that the sensors, environments, and platform setups are different between the KITTI odometry datasets to others, which thoroughly tests the generalization ability of the approaches. 

\begin{table}
	\caption{Registration results on multiple datasets using RANSAC- 50k. The results in brackets on the KITTI dataset are those reported in the original paper evaluated under bad ground truth poses. We fix it and also report the new results. The best results are highlighted in bold, and the second bests are marked with underlines. All the models are only trained on the KITTI dataset.}
	\centering
	\scriptsize	
	\setlength{\tabcolsep}{4pt}
%	\renewcommand\arraystretch{1.1}
	
% 		 		\scriptsize
	\begin{tabular}{l|c|c|c|c|c}
		\toprule
		%		\hline
		&KITTI     &KITTI-360   & Apollo  		&Mulran 	&Campus\\
%		\# Number of pairs	&555  		&6238 		&4250 			&3363		&171	\\
		%		\hline
%		\midrule
		% 		\multirow{2}*{Sensors} &\multirow{2}*{Velodyne 64}  &\multirow{2}*{Velodyne 64}  &\multirow{2}*{Velodyne 64} & Ouster 64	&\multirow{2}*{Velodyne 16}	\\
		% 		&   &   &  & (partially occluded)	&		\\
		
%		\multirow{1}*{Sensors} &\multirow{1}*{Velodyne 64}  &\multirow{1}*{Velodyne 64}  &\multirow{1}*{Velodyne 64} & Ouster 64	&\multirow{1}*{Velodyne 16}	\\
		%		&   &   &  & (partially occluded)	&		\\
		
		\midrule
		\multicolumn{6}{c}{\textit{Registration Recall}~(\%)}\\
		\midrule
		Predator~\cite{huang2021cvpr}	
		&\textbf{99.82} 		&99.50 			&\underline{99.27}  	&53.02 	&9.94\\ 
		CofiNet~\cite{yu2021nips}		
		&\textbf{99.82}			&99.62 			&\textbf{100}  		&80.79 		&36.84 \\ 
		NgeNet~\cite{zhu2022arxiv}		
		&\textbf{99.82}	    	&\textbf{99.94} 	&\textbf{100}  		&\underline{82.96} &\underline{81.29}\\ 
		Geotransformer~\cite{qin2022cvpr}
		&\textbf{99.82} 		&99.86 		&\textbf{100}  		&75.68 &71.93\\
		\name~(\textit{ours})	
		&\textbf{99.82}  &\underline{99.89}  &\textbf{100} &\textbf{87.09} &\textbf{96.49}  \\
		% 		\midrule
		% 		HRegNet~\cite{lu2021iccv}		
		% 		&96.76 		&20.39 			&9.39  	& -	   	&8.19\\
		% 		Geotransformer~(LGR)~\cite{qin2022cvpr}	
		% 		&\textbf{99.82} &99.86 		&\textbf{100}  		&72.91 &60.23\\
		% 		\name~(\textit{ours}, LGR)	
		% 		& \textbf{99.82} &\underline{99.90} &\textbf{100} &\underline{83.68} &\underline{89.47}  \\

		\midrule
		\multicolumn{6}{c}{\textit{Relative Rotation Error}~($^\circ$)}\\
		\midrule
		Predator~\cite{huang2021cvpr}	
		&0.25~(0.27) 		&0.29 			&0.21  		&1.03 	&1.94\\ 
		CofiNet~\cite{yu2021nips}		
		&0.37~(0.41)		&0.44 			&0.18  		&0.52 	&1.81\\ 
		NgeNet~\cite{zhu2022arxiv}		
		&0.26~(0.30)	    &0.30 			&0.18  		&\underline{0.35} 	&1.01\\ 
		Geotransformer~\cite{qin2022cvpr}
		&\underline{0.22}~(0.24) 	&\underline{0.28}	&\underline{0.12}  		&\textbf{0.30} 	&\underline{0.97}\\
		\name~(\textit{ours})	
		&\textbf{0.18} &\textbf{0.25} &\textbf{0.10} &{0.45} 	&\textbf{0.69}  \\
		% 		\midrule
		% 		HRegNet~\cite{lu2021iccv}		
		% 		&1.04 		&2.18 			&2.14  		& - 	&2.72\\
		% 		Geotransformer~(LGR)~\cite{qin2022cvpr}	
		% 		&0.31~0.31) 	&0.36 		&0.29  		&0.49 	&0.97\\
		% 		\name~(\textit{ours}, LGR)	
		% 		&{0.27} &{0.35} &{0.29} &{0.48} &\underline{0.89}  \\

		\midrule
		\multicolumn{6}{c}{\textit{Relative Translation Error}~(cm)}\\
		\midrule
		Predator~\cite{huang2021cvpr}	
		&\underline{5.8}~(6.8)		&\underline{7.2} 			&7.8  		&30.4 	&53.9\\ 
		CofiNet~\cite{yu2021nips}		
		&8.2~(8.5)		&10.1 			&6.7  		&17.3 	&38.6\\ 
		NgeNet~\cite{zhu2022arxiv}		
		&6.1~(7.4)	    &7.5			&\underline{5.9}  		&\textbf{9.2}   &\underline{13.6}\\ 
		Geotransformer~\cite{qin2022cvpr}
		&6.7~(7.4)		&8.1 			&{6.1}  		&\underline{12.0} 	&18.4\\
		\name~(\textit{ours})	
		&\textbf{5.3} &\textbf{7.0} &\textbf{4.6} &{14.4} &\textbf{12.7}\\
		% 		\midrule
		% 		HRegNet~\cite{lu2021iccv}		
		% 		&6.7		&112.8 		&116.7  	& - 	&98.3\\
		% 		Geotransformer~(LGR)~\cite{qin2022cvpr}
		% 		&5.5~(6.0) &\underline{6.9} 	&5.1  		&9.7 	&18.0\\
		% 		\name~(\textit{ours}, LGR)	
		% 		&\textbf{3.9} &\textbf{5.9} &\textbf{3.5} &\underline{9.3} &{15.3}  \\
		
		\bottomrule	
	\end{tabular}
		\vspace{-0.2cm}
	\label{tab:Registration}
\end{table}

%\vspace{-0.2cm}
\subsection{Correspondence Matching Performance}
\label{sec:corres}
We first evaluate the correspondence matching performance. 
Following~\cite{ qin2022cvpr}, we use two metrics to evaluate the matching performance: i) Inlier Ratio~(IR), the ratio of correct correspondences with residuals below a certain threshold, e.g., 0.6~m, after applying the ground truth transformation, and ii) Feature Match Recall~(FMR), the fraction of point cloud pairs with inlier ratio above a threshold, e.g., 5$\%$.

We compare the results of our method with the recent state-of-the-art methods: Predator~\cite{huang2021cvpr}, CofiNet~\cite{yu2021nips}, NgeNet~\cite{zhu2022arxiv}, and Geotransformer~\cite{qin2022cvpr}. 
We report the results in \tabref{tab:correspondences} with different numbers of samples. The sampling strategy is slightly different for different methods. 
For Predator and NgeNet, we use the default setting that samples points with probability proportional to the estimated scores.
As for our approach, CofiNet, and Geotransformer, because they directly output point correspondences without interest point sampling, we pick the top-$k$ correspondences according to fine level soft assignment scores.
Note that, the result of our approach reported on Mulran dataset is the result after removing the \mname{}, as we find that \mname{} is hard to generalize to partially occluded point clouds. 
For a fair comparison, we do not retrain the modified model. To our surprise, \name{} still achieves the best FMR and the second-best IR on Mulran, as shown in \tabref{tab:correspondences}. 
\name{} performs even better on other benchmarks, achieving the best FMR and IR. Notably, \name{} exceeds the baseline by a large margin of about 10\%-15\% in IR.
Interestingly, the performance of the keypoint-based methods and the methods that follow a coarse-to-fine manner present different trends when the number of samples becomes smaller. Keypoint-based methods, i.e., Predator and NgeNet, show a downward trend, while coarse-to-fine methods, i.e., CofiNet, Geotransformer, and ours, show an upward trend.
The reason is that when the number of samples becomes smaller, the keypoint-based method becomes more difficult to sample the points in the overlap region, which is underwritten by the reliable patch correspondences for coarse-to-fine methods.

\begin{figure}[t] 
	\centering
%	\DIFaddbeginFL \includegraphics[width=0.99\linewidth]{pics/recall.png} \DIFaddendFL
	\includegraphics[width=0.99\linewidth]{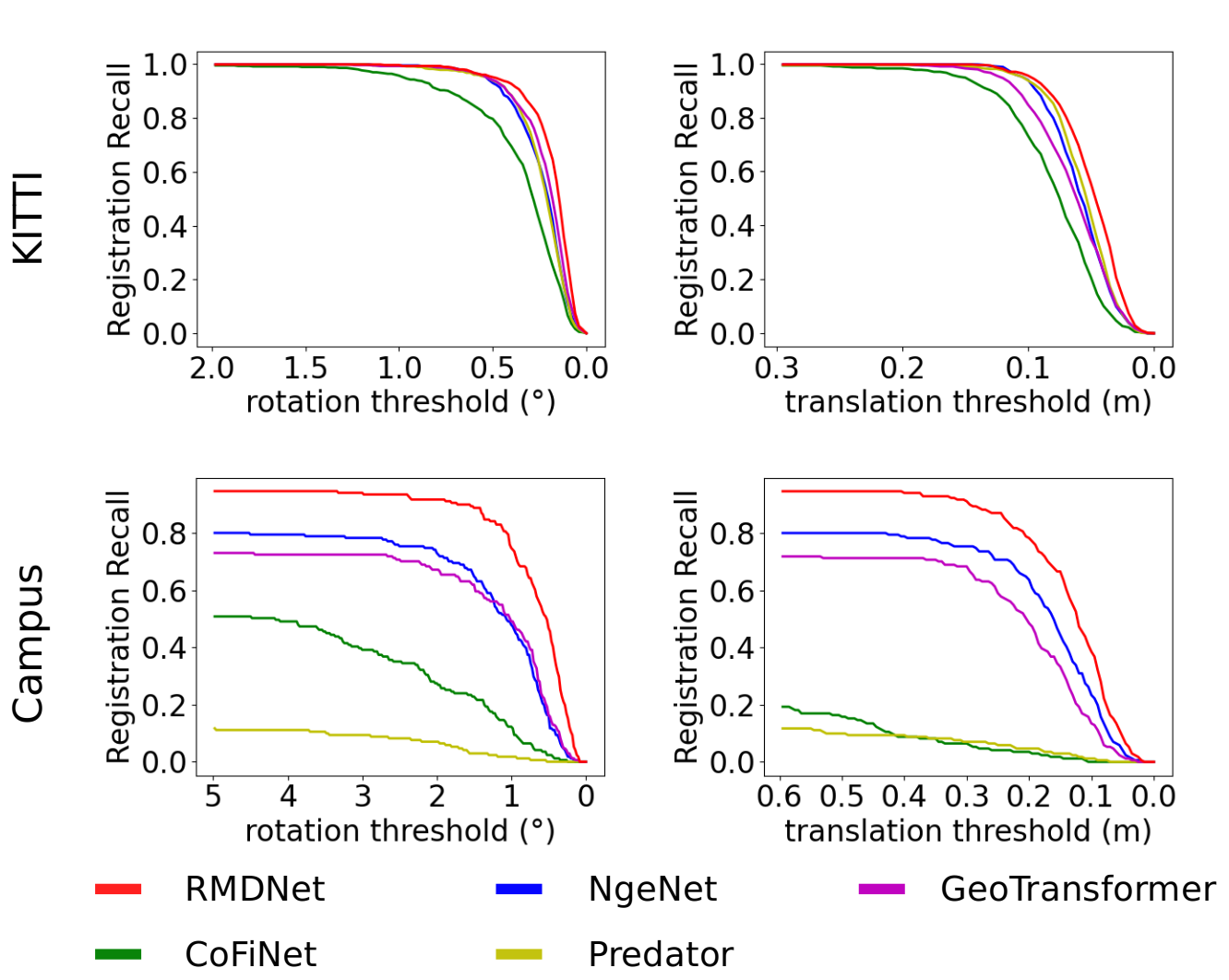}
	\caption{\diff{The registration recall under different thresholds. When changing one of the thresholds, we fix the other one to its default number, i.e., 5$^\circ$ for the rotation threshold and 2~m for the translation threshold.} }
	\label{fig:recall_curve} 
		\vspace{-0.2cm}
\end{figure}
 
 \begin{table}
 	\caption{Registration results on multiple datasets without RANSAC. All the models are only trained on the KITTI dataset.}
 	\centering
 	\scriptsize
 	\setlength{\tabcolsep}{4pt}
 	% 		\footnotesize
 	\begin{tabular}{l|c|c|c|c|c}
 		\toprule
 		%		\hline
 		&KITTI     &KITTI-360   & Apollo  		&Mulran 	&Campus\\
 %		\# Number of pairs	&555  		&6238 		&4250 			&3363		&171	\\
 		%		\hline
 %		\midrule
 % 		\multirow{2}*{Sensors} &\multirow{2}*{Velodyne 64}  &\multirow{2}*{Velodyne 64}  &\multirow{2}*{Velodyne 64} & Ouster 64	&\multirow{2}*{Velodyne 16}	\\
 % 		&   &   &  & (partially occluded)	&		\\
 %		\multirow{1}*{Sensors} &\multirow{1}*{Velodyne 64}  &\multirow{1}*{Velodyne 64}  &\multirow{1}*{Velodyne 64} & Ouster 64	&\multirow{1}*{Velodyne 16}	\\

 		\midrule
 		\multicolumn{6}{c}{\textit{Registration Recall}~(\%)}\\
 		\midrule
 		
 %		Best RANSAC result		
 %		&\textbf{99.82} 		&\textbf{99.94} 			&\textbf{100}  	& \textbf{87.09}	   	&\textbf{92.98}\\
 %		\midrule
 		HRegNet~\cite{lu2021iccv}		
 		&{96.76} 		&20.39 			&9.39  	& -	   	&8.19\\
 		Geotransformer~(LGR)
 		&\textbf{99.82} &99.86 		&\textbf{100}  		&72.91 &60.23\\
 		\name~(\textit{ours}, LGR)	
 		& \textbf{99.82} &\textbf{99.90} &\textbf{100} &\textbf{83.68} &\textbf{86.55} \\

 		\midrule
 		\multicolumn{6}{c}{\textit{Relative Rotation Error}~($^\circ$)}\\
 		\midrule
 		
 %		Best RANSAC result		
 %		&\textbf{0.18} 		&\textbf{0.25} 			&\textbf{0.10}  	& \textbf{0.30}	   	&\textbf{0.76}\\
 %		\midrule
 		HRegNet~\cite{lu2021iccv}		
 		&1.04 		&2.18 			&2.14  		& - 	&2.72\\
 		Geotransformer~(LGR)	
 		&0.31~(0.31) 	&0.36 		&{0.29}  		&0.49 	&0.97\\
 		\name~(\textit{ours}, LGR)	
 		&\textbf{0.27} &\textbf{0.35} &\textbf{0.29} &\textbf{0.48} &\textbf{0.83}  \\

 		\midrule
 		\multicolumn{6}{c}{\textit{Relative Translation Error}~(cm)}\\
 		\midrule
 		
 %		Best RANSAC result		
 %		&\underline{5.3} 		&{7.0}		&\underline{4.6}  	& \textbf{9.2}	   	&\textbf{13.1}\\
 %		\midrule
 		HRegNet~\cite{lu2021iccv}		
 		&6.7		&112.8 		&116.7  	& - 				&98.3\\
 		Geotransformer~(LGR)
 		&5.5~(6.0) &{6.9} 	&5.1  		&9.7 	&18.0\\
 		\name~(\textit{ours}, LGR)	
 		&\textbf{3.9} &\textbf{5.9} &\textbf{3.5} &\textbf{9.3} &\textbf{16.1}  \\

 		\bottomrule	
 	\end{tabular}
 %		\vspace{-0.5cm}
 	\label{tab:Registration_lgr}
 \end{table}
% \begin{table}
% 	\caption{Correspondence results on multi-modality datasets.}
% 	\centering
% 	\scriptsize
% 	\setlength\tabcolsep{3pt}
% 	%	\footnotesize
%% 	\resizebox{\linewidth}{!}
% 	{\begin{tabular}{l|ccccc|cccc}
% 		\toprule
% 		& \multicolumn{5}{c|}{Mulran} 	&   \multicolumn{4}{c}{Campus}\\
% 		
% 		\# Samples &all&5000&2000&1000&250 	&all&2000&1000&250\\
% 		
% 		\midrule
% 		\multicolumn{10}{c}{\textit{Feature Match Recall}~(\%)}\\
% 		\midrule
% 		
% 		Geotransformer~\cite{qin2022cvpr}
% 		& \textbf{}&\textbf{88.28}&\textbf{}&91.05&91.67 		&  &\underline{98.25}&\textbf{98.25}&97.66		\\
% 		\name~(\textit{ours})  &\textbf{}&\textbf{98.31}&\textbf{}&98.72&98.81  &\textbf{100}&\textbf{100}&\textbf{100}&\textbf{100}  \\
% 		
% 		\midrule
% 		\multicolumn{10}{c}{\textit{Inlier Ratio}~(\%)}\\
% 		\midrule
% 		
% 		Geotransformer~\cite{qin2022cvpr}
% 		&& \textbf{33.6} &&\textbf{43.7}&\underline{46.3}	&&\underline{19.0}&\underline{21.7}&\underline{24.9}\\
% 		\name~(\textit{ours})	   &&\underline{31.4}&&\underline{42.6}&\textbf{51.1} 	&&\textbf{34.9}&\textbf{36.8}&\textbf{41.5}  \\
% 		
% 		\bottomrule	
% 	\end{tabular}}
% 	
% 	%	\vspace{-0.2cm}
% 	\label{tab:comparetogeo}
% \end{table}

 \begin{figure*}[t] 
 	\centering
 	\includegraphics[width=0.9\linewidth]{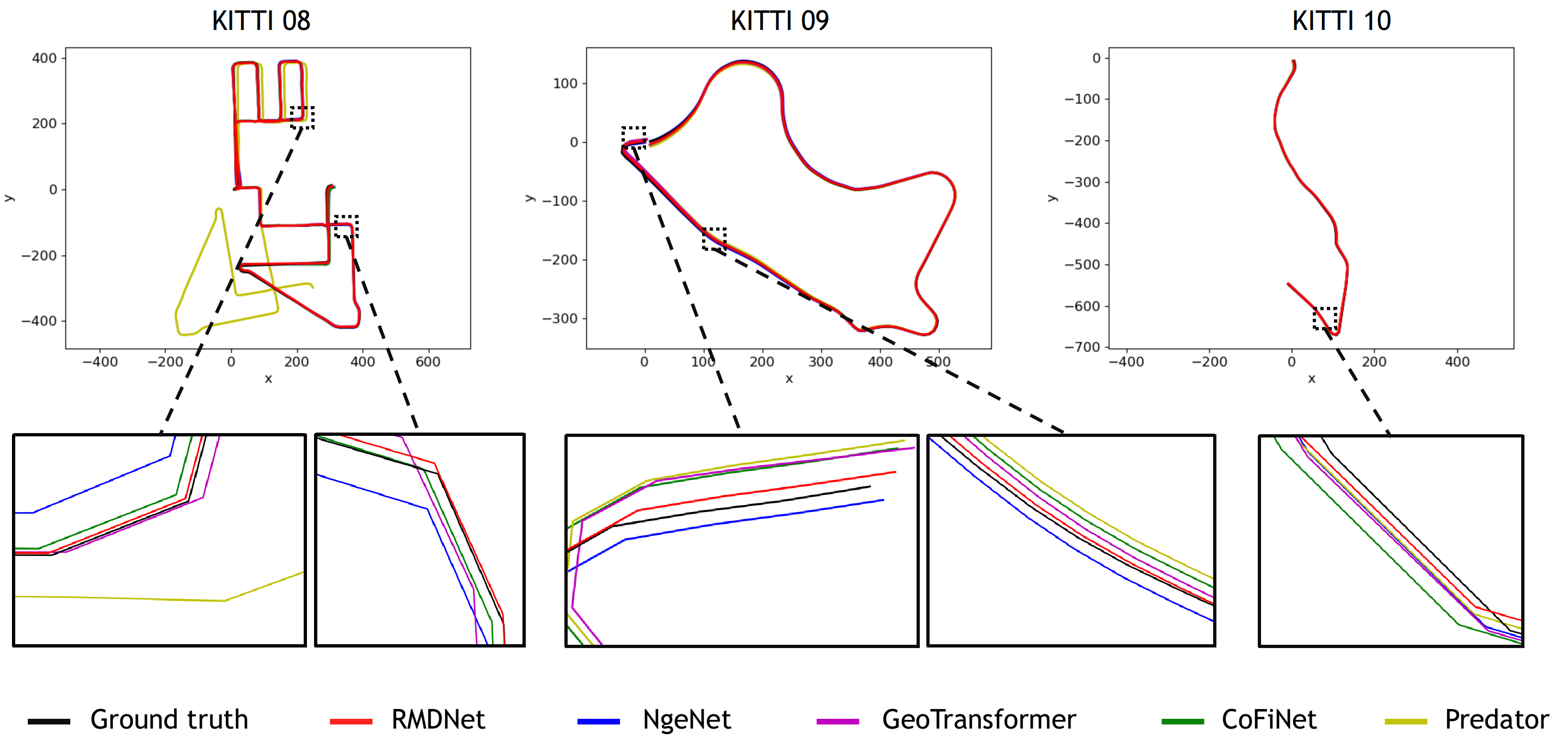}
 	\caption{Trajectory estimation results on KITTI dataset using the LiDAR pairs 10~m away.}
 	\label{fig:traj} 
 		\vspace{-0.2cm}
 \end{figure*}

\vspace{-0.2cm} 	
\subsection{Point Cloud Registration Performance}
The second experiment evaluates the point cloud registration results and supports our claim that our method outperforms the state-of-the-art method in point cloud registration.

We also follow~\cite{ qin2022cvpr} using three other metrics to evaluate the registration performance: i) Relative Translation Error~(RTE), the Euclidean distance between estimated and ground truth translation vectors, ii) Relative Rotation Error~(RRE), the geodesic distance between estimated and ground truth rotation matrices, and iii) Registration Recall~(RR), the fraction of scan pairs whose RRE and RTE are below certain thresholds, e.g., 5$^\circ$ and 2~m.

%We report the result of \name{} with \mname{} removed on Mulran.
We compare the results of our method with the recent RANSAC-based state-of-the-art methods: Predator~\cite{huang2021cvpr}, CofiNet~\cite{yu2021nips}, NgeNet~\cite{zhu2022arxiv}, and Geotransformer~\cite{qin2022cvpr} in \tabref{tab:Registration}.
There is an error in the evaluation codes of these methods using KITTI ground-truth except NgeNet.
We fix the error and report the results before the fix as reported in the original papers and new results after the fix.
As can be seen, Our \name{} achieves the best RR on Mulran and outperforms all the baselines on all other datasets. 
%Notably, our \name{} shows remarkable performance when generalizing to the campus dataset by boosting baselines by a large margin on all metrics.
\diff{Especially on the campus dataset, \name{} outperforms baseline methods with a large margin on all metrics. 
We further evaluate RR for all the methods at different RRE and RTE thresholds on the KITTI and Campus datasets (see \figref{fig:recall_curve}). Our RMDNet exhibits higher registration recall at all thresholds. Especially, RMDNet exceeds baseline methods by a large margin when generalizing to the campus dataset.}

\begin{table}
	\caption{Absolute pose error on sequences 08-10 of KITTI dataset. Best performance is highlighted in bold while the second best is marked with an underline.}
	\centering
	\scriptsize	
	\setlength{\tabcolsep}{2pt}
	%	\renewcommand\arraystretch{1.1}
	
	% 		 		\scriptsize
	\begin{tabular}{l|c|c|c|c|c|c}
		\toprule
		%		\hline
		&Rot.     &Rot.  &Rot.  & Trans.  	&Trans.	&Trans. 	\\
		&RMSE~($^\circ$)     &MAE~($^\circ$) &STD~($^\circ$)  & RMSE(cm)   &MAE(cm)  &STD(cm)	\\
		
		\midrule
		\multicolumn{7}{c}{\textit{KITTI Sequence 08}}\\
		\midrule
		Predator~\cite{huang2021cvpr}	
		&104.7 		&39.77 		&{8.13}  	&10469.6&4228.3&4319.7	\\ 
		CofiNet~\cite{yu2021nips}		
		&8.80		&4.11 	&1.77		&879.8  	&335.1 &381.8	\\ 
		NgeNet~\cite{zhu2022arxiv}		
		&{6.48}	    &3.17	&0.91  		&647.8 &273.4 &255.2\\ 
		Geotransformer~\cite{qin2022cvpr}
		&\underline{4.18} 		&\underline{2.01} 	&\textbf{0.60}  		&\underline{418.3} &\underline{171.2}&\underline{170.3}\\
		\name~(\textit{ours})	
		&\textbf{3.23}  &\textbf{1.58} &\underline{0.68} &\textbf{323.0} &\textbf{139.3}&\textbf{124.0} \\

		\midrule
		\multicolumn{7}{c}{\textit{KITTI Sequence 09}}\\
		\midrule
		Predator~\cite{huang2021cvpr}	
		&{4.14}		&1.99 		&1.12&413.5	&174.9 		&162.5 	\\ 
		CofiNet~\cite{yu2021nips}		
		&3.10		&2.35 			&1.61  &309.8&121.5		&131.3 	\\ 
		NgeNet~\cite{zhu2022arxiv}		
		&4.67	    &2.54	&0.76&466.7		&173.0  		&206.6   \\ 
		Geotransformer~\cite{qin2022cvpr}
		&\underline{2.84}		&\textbf{1.37} 			&\underline{0.76} &\underline{283.6}&\underline{103.6} 		&\underline{126.8} 	\\
		\name~(\textit{ours})	
		&\textbf{1.91} &\underline{1.43} &\textbf{0.67} &\textbf{190.9} &\textbf{76.9}&\textbf{79.0} \\

		\midrule
		\multicolumn{7}{c}{\textit{KITTI Sequence 10}}\\
		\midrule
		Predator~\cite{huang2021cvpr}	
		&3.5		&2.55 			&1.37  &349.7&136.6		&148.7 	\\ 
		CofiNet~\cite{yu2021nips}		
		&3.2		&2.9 			&1.6  	&319.9&134.6	&126.5 	\\ 
		NgeNet~\cite{zhu2022arxiv}		
		&2.97	    &2.2 	&1.08&296.8	&106.7  		&134.1 	\\ 
		Geotransformer~\cite{qin2022cvpr}
		&\underline{1.32}	&\underline{1.1}	&\textbf{0.35}  		&\underline{132.5} &\underline{64.5}&\underline{41.2}	\\
		\name~(\textit{ours})	
		&\textbf{1.26} &\textbf{1.09} &\underline{0.39} &\textbf{126.3} &\textbf{60.9}&\textbf{40.0}\\

		\bottomrule	
	\end{tabular}
	%		\vspace{-0.2cm}
	\label{tab:ate}
\end{table}

%\begin{figure*}[t] 
%	\centering
%	\includegraphics[width=0.99\linewidth]{pics/qualitative}
%	\caption{Qualitative results of the learned superpoints. The offsets from the initial uniformly sampling nodes to the learned superpoints are denoted as \textcolor{red}{red line}.}
%	\label{fig:qualitative} 
%\end{figure*}

\begin{figure*}[t] 
	\centering
	\includegraphics[width=0.99\linewidth]{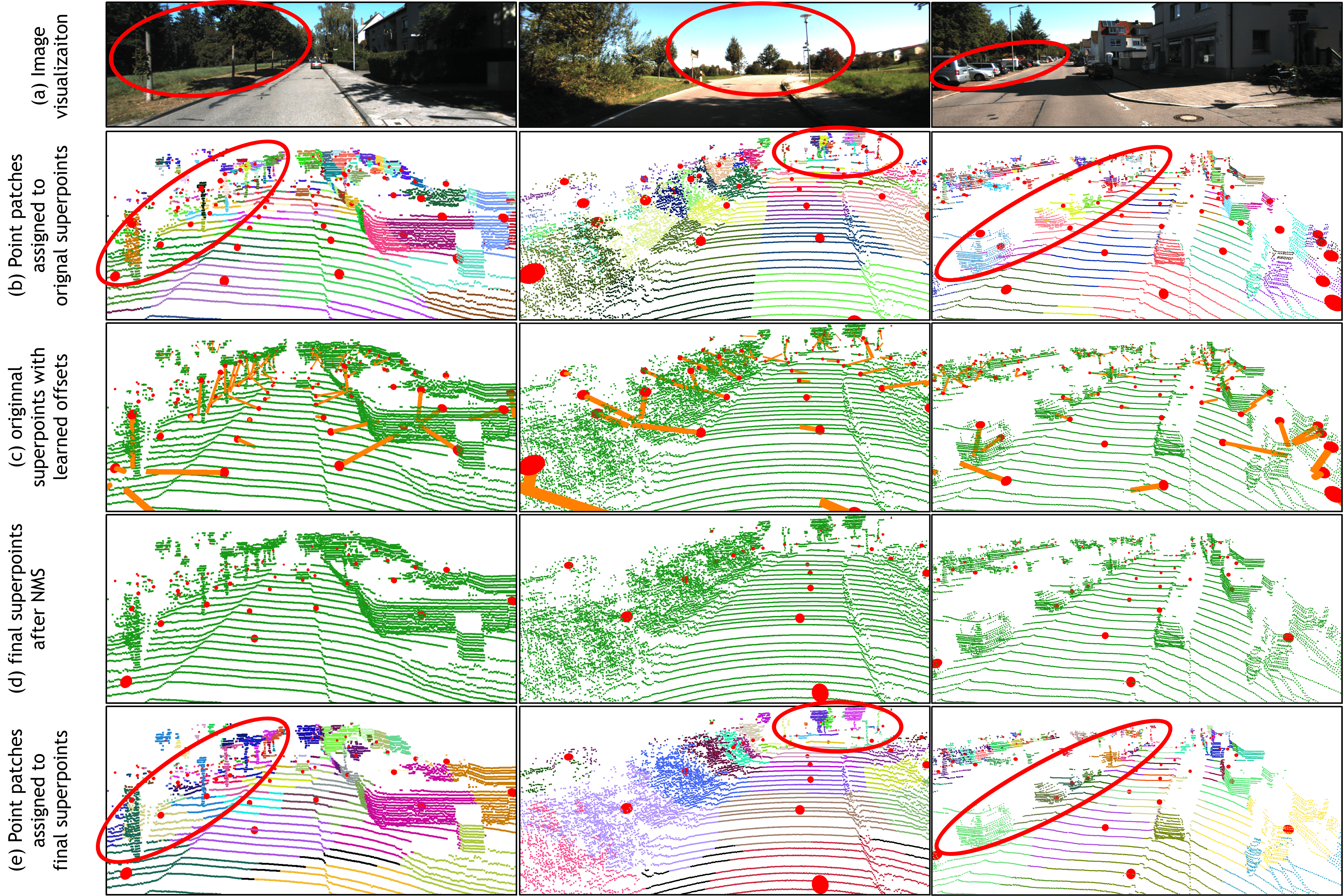}
	\caption{Visualizations of our superpoint detection and point patch grouping. (a) shows the corresponding images of the environment only for reference. (b) shows the uniformly sampled superpoints (\textcolor{red}{red dots}) with their point patches grouped by point-to-node strategy. We assign each point patch a random color for visualization. (c) shows the uniformly sampled superpoints with learned offsets (\textcolor{orange}{orange lines}). (d) shows the final superpoints after offsetting and filtering. (e) shows the point patches grouped by the final superpoints. Each point patch is assigned a random color.
	Uniformly sampled superpoints can easily separate a single object into several parts, as seen in (b). This poses challenges for matching. Our superpoint detection module learns a pattern that brings the superpoints close to a geometrically significant region together without using any semantic information. This leads to a more reasonable point patch grouping which benefits point matching.}
	\label{fig:vote_vis} 
	%		\vspace{-0.2cm}
\end{figure*}

We also compare our method to state-of-the-art RANSAC-free methods using Local-to-Global Registration~(LGR)~\cite{qin2022cvpr}: HRegNet~\cite{lu2021iccv} and Geotransformer~\cite{qin2022cvpr} in \tabref{tab:Registration_lgr}.
LGR is specifically proposed for superpoint-based approaches~\cite{qin2022cvpr}.
It calculates poses by performing weighted SVD on dense point correspondences of each patch and chooses one that admits the most inlier matches, which greatly reduces the computation time by limiting the iterations to $|\mathcal{M}|$.
When using LGR, our method attains remarkable results for translation estimation and surpasses the best RANSAC-based results by about 1~cm on KITTI, KITTI-360, and Apollo datasets.
In most cases, our method is the best among other RANSAC-free methods.

The above experiments evaluate the performance of all the methods in relative pose estimation. We further evaluate the absolute pose error of all the methods. 
We still use the LiDAR pairs 10~m away for transformation estimation and chain these transformations to obtain the trajectory. \figref{fig:traj} shows the trajectories estimated by different methods on the three test sequences of the KITTI dataset. We calculate the root mean squared error~(RMSE) and mean absolute error~(MAE) of each estimated trajectory against the ground truth trajectory listed in \tabref{tab:ate}. As can be seen, our RMDNet achieves overall the best performance. 

So far, we have demonstrated the superior capability of our RMDNet for point cloud registration in terms of both relative and absolute pose errors. The superiorities of our method lay both in its high accuracy on the datasets similar to the training set (KITTI-360 and Apollo) and in its robustness while generalizing to the datasets under totally different sensor configurations (Mulran and Campus). The key to the robustness and accuracy are the two main modules of the method, i.e., 3D-RoFormer and superpoint detection module. The 3D-RoFormer injects powerful feature representation capability into the network by encoding relative position information in a lightweight manner,  while the superpoint detection module generates reliable superpoints that boost the matching and registering significantly. 
%Following, we provide more visual examples to show why the detected superpoints benefit the matching and the registering.

\begin{figure*}[t] 
	\centering
	\includegraphics[width=0.99\linewidth]{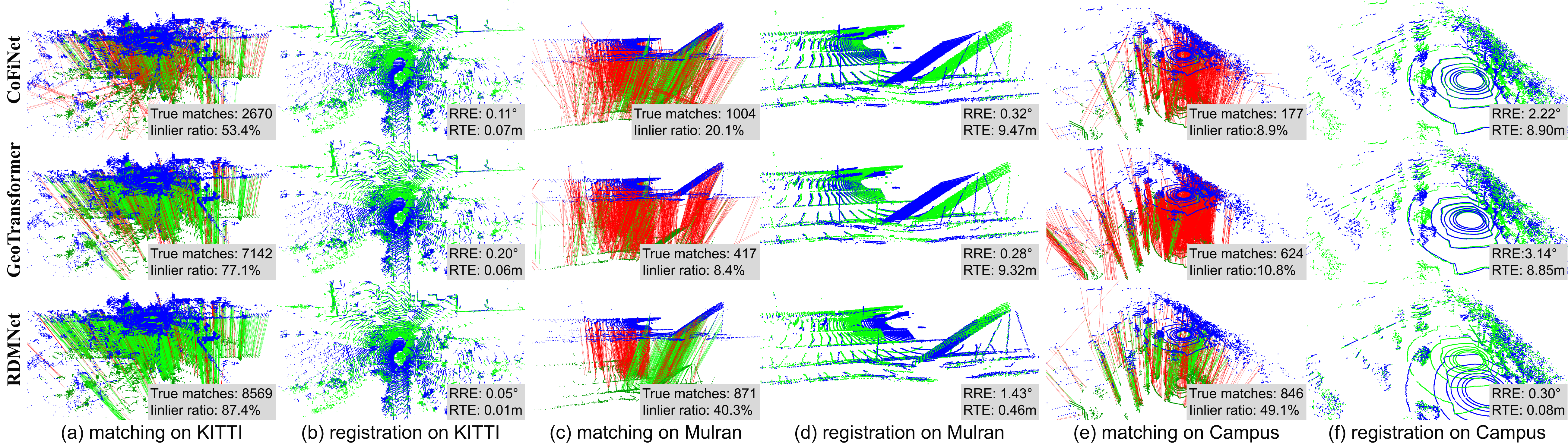}
	\caption{ Matching and registration results of our RDMNet compared to recent advances CoFiNet~\cite{yu2021nips} and GeoTransformer\cite{qin2022cvpr}. In (a), (c), and (e), we visualize the matching on three datasets. Our RMDNet finds more inlier matches on salient regions (e.g., the curbs and the shrubs) and reject the outlier matches between similar flat patches. (b), (d), and (f) show our RMDNet achieves more accurate and robust registration compared to baseline methods.}
	\label{fig:match_vis} 
	%		\vspace{-0.2cm}
\end{figure*}

%Here we briefly review the process of superpoint generation, we first sample uniformly distributed sample points from the point cloud, then learn the offset from these nodes to the proposals, and remove the close proposals to obtain the final superpoints.
%There are two properties of our learned offsets: i) only the nodes near the geometrically significant region generate recognizable offsets; ii) The learned offsets bring the nodes near a geometrically significant region closer to each other.

%%%%%%%%%%%%%%%%%%%%%%%%%%
%\subsection{Qualitative Result}
%In this section, we provide more qualitative results in \figref{fig:qualitative} to show the characteristics of the learned superpoints, and explain why they can improve the quality of the correspondences.
%
%There are two properties of our learned offsets: i) only the nodes near the geometrically significant region generate recognizable offsets; ii) The learned offsets bring the nodes near a geometrically significant region closer to each other.
%
%These two properties bring two benefits for point-level matches, respectively. First, the resulting nodes are roughly distributed in the whole point cloud and not concentrated in some specific regions, ensuring that most of the points will be included in the vicinity of superpoints, increasing the density of the final correspondences. Second, after filtering the close superpoints, a geometrically significant region is rarely shared by multiple superpoints, making the points in the region easier to match as a whole.

%%%%%%%%%%%%%%%%%%%%%
\subsection{Qualitative Results}
To provide more insights into the proposed \name{}, we visualize the superpoints and the corresponding point patches learned by our method in~\figref{fig:vote_vis}. As can be seen, compared to the vanilla uniformly distributed superpoints (second row), our method can extract the superpoints close to their nearest geometrically significant region, such as the curbs and objects, as shown from the third row to the fourth row. Based on that, our method groups the dense point patches more meaningfully, where points on a single object are gathered in the same patch (fifth row), which is important for finding accurate correspondence and obtaining good registration results. 

\figref{fig:match_vis} provides the matching and registration results of our method compared to Geotransformer~\cite{qin2022cvpr} and CoFiNet~\cite{yu2021nips}. It shows that our RDMNet finds more inlier matches on salient regions, rejects outlier matches between similar flat areas, and performs robust and accurate registration.

%%%%%%%%%%%%%%%%%%%%%
\diff{\subsection{Robustness Tests}
\begin{figure}[t] 
  	\centering
%  	\DIFaddbeginFL \includegraphics[width=0.99\linewidth]{pics/robust.png} \DIFaddendFL
	\includegraphics[width=0.99\linewidth]{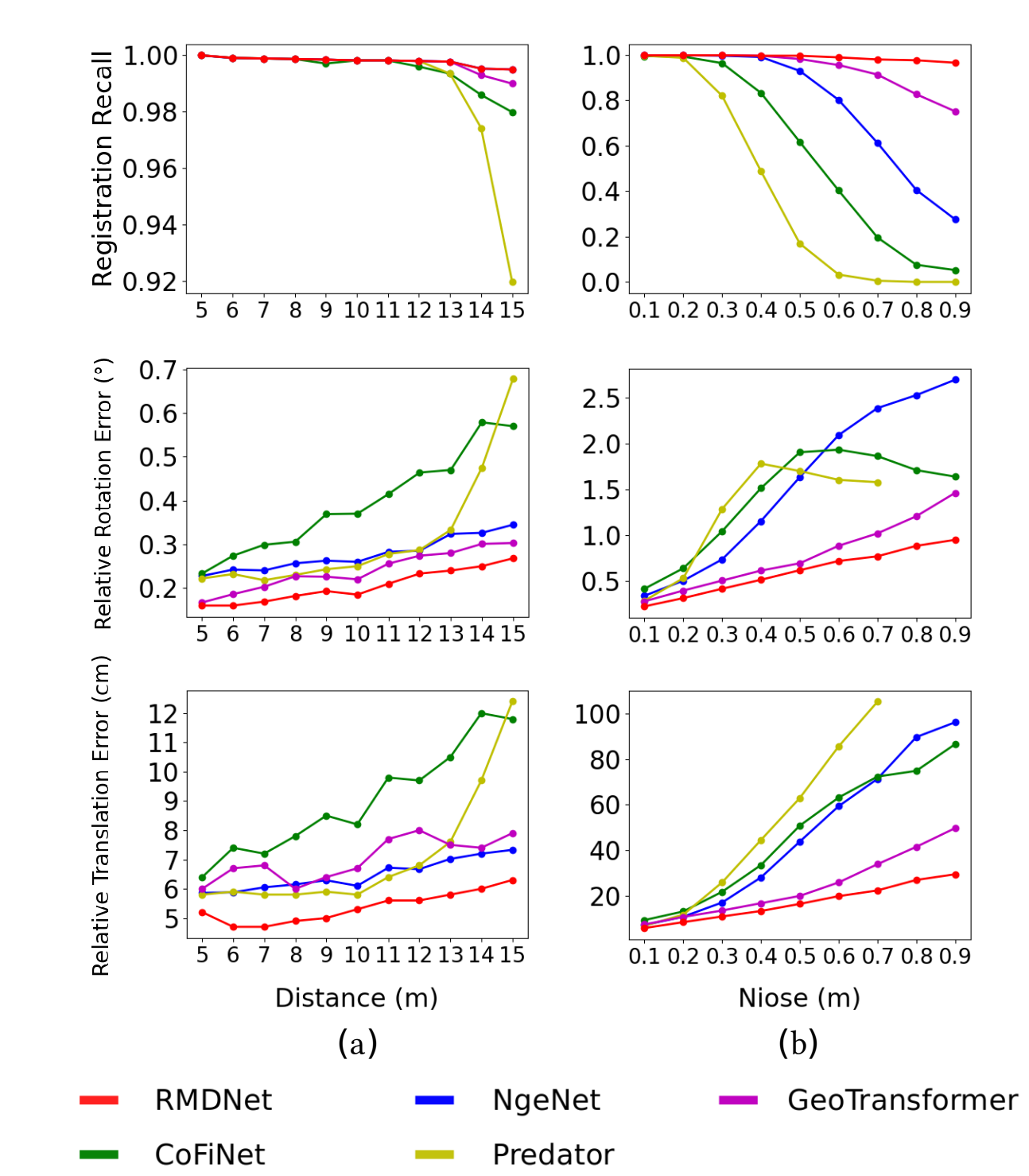}
  	\caption{\diff{Registration robustness tests in terms of (a) pair-wise distance and (b) noise level on the KITTI datasets.}}
  	\label{fig:robust} 
  		\vspace{-0.2cm}
  \end{figure}
We conduct registration robustness tests regarding different overlap ratios and noise levels on the KITTI datasets. 
We generate the testing datasets with varying overlap ratios using the LiDAR pairs at different distances. See \figref{fig:robust}a, in terms of RR, RRE, and RTE, \name{} achieves the best registration performance for paired point clouds with varying overlap ratios.
For evaluation under different noise levels, we add zero-mean Gaussian noise with $\sigma$ standard deviation to the point coordinates. See \figref{fig:robust}b, \name{} obtains the more accurate and robust registration than competitor algorithms at all noise levels. Note that \name{} shows the superior robustness that maintains an extremely high registration recall of $96.58\%$ compared to other baselines at a high noise level of $0.9$~m.}

%%%%%%%%%%%%%%%%%%%%%%%%
\subsection{Ablation Study}

We conduct ablation studies on KITTI and Apollo datasets to better understand the effectiveness of each module in the proposed \name{}, and show that the full \name{} is the best setup. 
We use the model trained with negative log-likelihood loss~\cite{qin2022cvpr,yu2021nips} using the vanilla transformer and without the \mname{} as the base model.
%We use the model using the vanilla transformer and without using \mname{} that trained with the negative log-likelihood loss as the base model.
 \tabref{tab:ablation} summarizes the point registration results of the ablation study. SDM refers to the \mname{}, and RoPE refers to the rotary position embedding.
%SDM represents the \mname{}. RoPE represents the results using vanilla transformer. ``w/o gap loss" represents the results trained with the negative log-likelihood loss used in ~\cite{qin2022cvpr,yu2021nips}. ``Full RDMNet" is the complete proposed \name{}. 
As can be seen, all modules of our method bring improvement in the point cloud registration individually. Combining all proposed modules, our \name{} performs the best.

We also provide a study on the effectiveness of our proposed 3D-RoFormer. We compare it with other existing transformers, including vanilla transformer~\cite{yu2021nips}, absolute position embedding~(APE)~\cite{shi2021ral}, and geometric embedding~(GEO)~\cite{qin2022cvpr}. We only change the transformer parts in our \name{} while keeping the rest parts the same and comparing the point cloud registration results. As shown in \tabref{tab:ablation_roformer}, using our proposed 3D-RoFormer, our \name{} achieves the best performance in all metrics on both the KITTI and Campus datasets.
 
\begin{table}
	\caption{Ablation study of individual modules.}
	\centering
	\scriptsize
	\setlength\tabcolsep{6pt}
	
	% 		\footnotesize
	\begin{tabular}{ccc|ccc|ccc}
		\toprule
		%		\hline
		\multirow{2}*{SDM}&\multirow{2}*{RoPE}&\multirow{2}*{gap loss} &\multicolumn{3}{c|}{KITTI}  		&\multicolumn{3}{c}{Apollo}\\
		&&&\multirow{1}*{RR}  &\multirow{1}*{RRE}  &\multirow{1}*{RTE} & RR	&\multirow{1}*{RRE} & RTE	\\
		
		\midrule
%		& &\checkmark &		& 			&  		& 	& 	& \\
%		w/o SDM		
		&\checkmark &\checkmark &99.46 		&0.20 			&5.8  	&99.34 	   	&0.16 &6.9\\
%		w/o RoPE
		\checkmark& &\checkmark &\textbf{99.82}	 	&0.20 			&5.5  		&98.49 &0.20&6.7\\
%		w/o gap loss		
		\checkmark&\checkmark& &\textbf{99.82} &\textbf{0.18}			&5.7  		&\textbf{100} 	&0.11  &5.0\\
%		Full \name{}		
		\checkmark&\checkmark&\checkmark &\textbf{99.82} 		&\textbf{0.18} 			&\textbf{5.3}  	& \textbf{100}	   	&\textbf{0.10}&\textbf{4.6}\\
		\bottomrule	
	\end{tabular}
	
%		\vspace{-0.2cm}
	\label{tab:ablation}
\end{table}

\begin{table}
	\caption{Ablation of 3D-RoFormer.}
	\centering
	\scriptsize
	\setlength\tabcolsep{6pt}
	
	% 		\footnotesize
	\begin{tabular}{l|ccc|ccc}
		\toprule
		%		\hline
		\multirow{2}*{Model} &\multicolumn{3}{c|}{KITTI}  		 &\multicolumn{3}{c}{Campus}\\
		&{RR}  &{RRE}  &{RTE} & RR	&{RRE} & RTE \\
		
		\midrule
		vanilla transformer		
		&\textbf{99.82} 		&0.32 			&7.3  	 &86.55 	&1.70	&26.6\\
		APE transformer
		&\textbf{99.82} &0.20 		&\textbf{5.3}  		 &91.23 	&0.97	&15.6\\
		GEO transformer
		&\textbf{99.82}  &0.24		&6.5 		&88.89 	&1.01	&20.2\\
		3D-RoFormer (Ours)	
		&\textbf{99.82} 		&\textbf{0.18} 			&\textbf{5.3}  		 &\textbf{96.49}	&\textbf{0.69}	&\textbf{12.7}\\
		\bottomrule	
	\end{tabular}
	
%		\vspace{-0.2cm}
	\label{tab:ablation_roformer}
\end{table}

%%%%%%%%%%%%%%%%%%%%%%%%%%%%%%%%%%%%%%%%%%%%%%%%%%%%%%%%%%%%%%%%%%%%%%%%%%%%%%%%
\subsection{Study on Runtime and Storage}
\label{sec:runtime} 
We measure the runtime and storage of the major module 3D-RoFormer for different numbers of input nodes per scan. The experiments are performed on an NVIDIA GeForce GTX 3090 GPU. As shown in~\tabref{tab:runtime}, the runtime and storage of 3D-RoFormer increase linearly as the number of input nodes increases, similar to the vanilla transformer and APE transformer, while less than the quadratic GEO transformer.

\begin{table}
	\caption{Runtime and storage of different transformer networks with different numbers of input nodes.}
	\centering
	\scriptsize
	\setlength\tabcolsep{6pt}
	
	% 		\footnotesize
	\begin{tabular}{l|ccc|ccc}
		\toprule
		%		\hline
		Model &\multicolumn{3}{c|}{Run time (ms)}	&\multicolumn{3}{c}{Storage (MB)}  \\
		\midrule
		Num. of input nodes&1000 &500  &100 & 1000	&500 & 100\\
		
		\midrule
		vanilla transformer		
		&{31} &24 &22&40& {23}	&6\\
		APE transformer
		&33 		&29  		 &24&46&24 &{6}\\
		GEO transformer
		&{58} 		&40	 &26 &{9784}&2318&6\\
		3D-RoFormer (Ours)	
		&{38} &{30}  	 & {28}	&{57} &{40}	&{6}\\
		\bottomrule	
	\end{tabular}
	
%		\vspace{-0.2cm}
	\label{tab:runtime}
\end{table}
%%%%%%%%%%%%%%%%%%%%%%%%%%%%%%%%%%%%%%%%%%%%%%%%%%%%%%%%%%%%%%%%%%%%%%%%%%%%%%%%
\section{Conclusion}
\label{sec:conclusion}

In this paper, we present \name{} that leverages a coarse-to-fine strategy to extract dense point correspondences for point cloud registration. We exploit insights from natural language processing and keypoint detection and design a novel transformation-invariant transformer named 3D-RoFormer. It learns to aggregate contextual and geometric information of the point clouds in a fast and lightweight way and extracts salient and compact superpoint pairs for point cloud registration. We evaluate and compare our approach on multiple datasets, including publicly available ones and our self-recorded dataset collected from different environments. \diff{Extensive experiments suggest that our approach outperforms the baseline methods in terms of both correspondence matching and point cloud registration with a strong generalization ability. In the future, we want to figure out why the voting module does not work well in the Mulran dataset and improve the voting module's generalization ability. We also want to explore the potential of \name{} to tackle the global localization problem. }

\small{
\bibliographystyle{unsrt}
\bibliography{reference,glorified}
}

\end{document}